\documentclass[10pt,twocolumn,letterpaper]{article}

\usepackage{iccv}
\usepackage{times}
\usepackage{epsfig}
\usepackage{graphicx}
\usepackage{amsmath}
\usepackage{amssymb}
\usepackage{url}
\usepackage{dsfont}
\usepackage{enumerate}
\usepackage{enumitem}
\usepackage{comment}
\usepackage{subfigure}
\usepackage[table, dvipsnames]{xcolor}
\usepackage{algorithm}
\usepackage{algpseudocode}
\usepackage[export]{adjustbox}
\usepackage{comment}
\usepackage{amssymb}
\usepackage[toc,page]{appendix}

\newcommand{\bx}[1]{\mathbf{x}_{#1}}
\newcommand{\by}[1]{\mathbf{y}_{#1}}

\newcommand{\lossfunc}[1]{L_{#1}}

\newcommand{\dropmask}{\mathbf{m}}
\newcommand{\argmax}{\mathop{\mathrm{argmax}}\limits}

\newcommand{\methodname}{DTA}
\newcommand{\sourcedomain}{\mathcal{S}}
\newcommand{\targetdomain}{\mathcal{T}}
\newcommand{\eadd}{EAdD}
\newcommand{\cadd}{CAdD}

\newcommand{\uppernet}{h_u}
\newcommand{\lowernet}{h_l}

\newcommand{\vmask}{\mathbf{v}}

\newcommand{\jacosum}{\mathbf{s}}
\newcommand{\absjacosum}{\mathbf{z}}
\newcommand{\sortedidx}{\mathbf{c}}
\newcommand{\zero}{\boldsymbol{0}}
\newcommand{\one}{\boldsymbol{1}}
\newcommand{\jacobian}{\mathbf{J}}

\newcommand\blfootnote[1]{%
	\begingroup
	\renewcommand\thefootnote{}\footnotetext{#1}%
	\addtocounter{footnote}{-1}%
	\endgroup
}

\usepackage[breaklinks=true,bookmarks=false,colorlinks]{hyperref}

\iccvfinalcopy 

\def\iccvPaperID{3788} 

\ificcvfinal\fi
\begin{document}

\title{Drop to Adapt: \\Learning Discriminative Features for Unsupervised Domain Adaptation}


\author{Seungmin Lee$^{\ast}$\\
	Seoul National Univ.\\
	\and
	Dongwan Kim$^{\ast}$\\
	Seoul National Univ.\\
	\and
	Namil Kim\\
	NAVER LABS\\
	\and
	Seong-Gyun Jeong\\
	CODE42.ai\\
}
\maketitle
\blfootnote{\hspace{-0.5cm} $^{\ast}$ denotes equal contribution.\\
This work was done while the authors were at NAVER LABS. Correspondence to \tt{\{profile2697, dongwan123\}@gmail.com}}

\begin{abstract}

Recent works on domain adaptation exploit adversarial training to obtain domain-invariant feature representations from the joint learning of feature extractor and domain discriminator networks. However, domain adversarial methods render suboptimal performances since they attempt to match the distributions among the domains without considering the task at hand. We propose Drop to Adapt (DTA), which leverages adversarial dropout to learn strongly discriminative features by enforcing the cluster assumption. Accordingly, we design objective functions to support robust domain adaptation. We demonstrate efficacy of the proposed method on various experiments and achieve consistent improvements in both image classification and semantic segmentation tasks. Our source code is available at \url{https://github.com/postBG/DTA.pytorch}.

\end{abstract}

\section{Introduction}\label{sec:intro}

\begin{figure}[t]
	\centering
	\footnotesize
	\begin{tabular}{cc}
		\includegraphics[width=4cm]{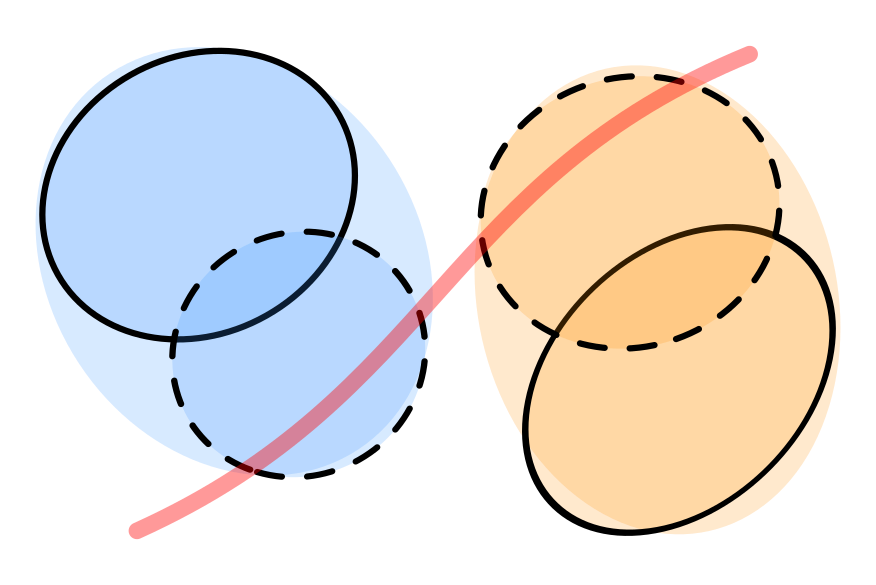}&
		\includegraphics[width=4cm]{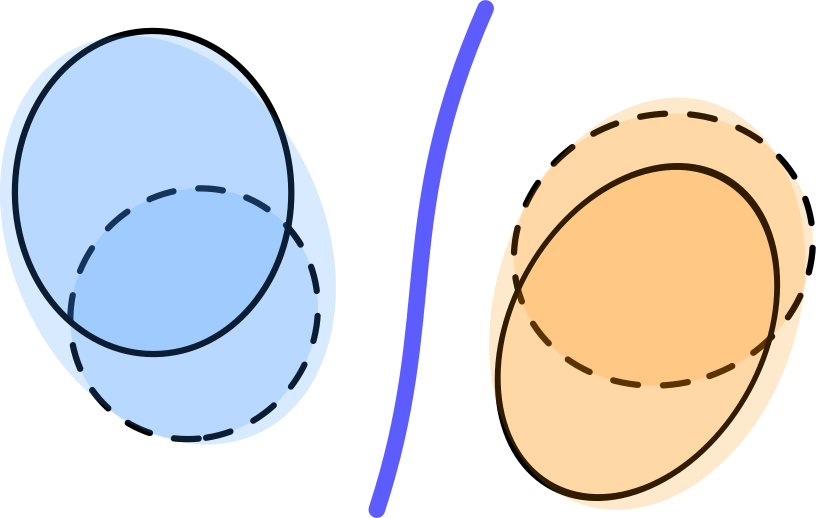}\\	
		(a) Before adaptation & (b) Adapted model\vspace{0.2cm}\\
		\includegraphics[width=4cm]{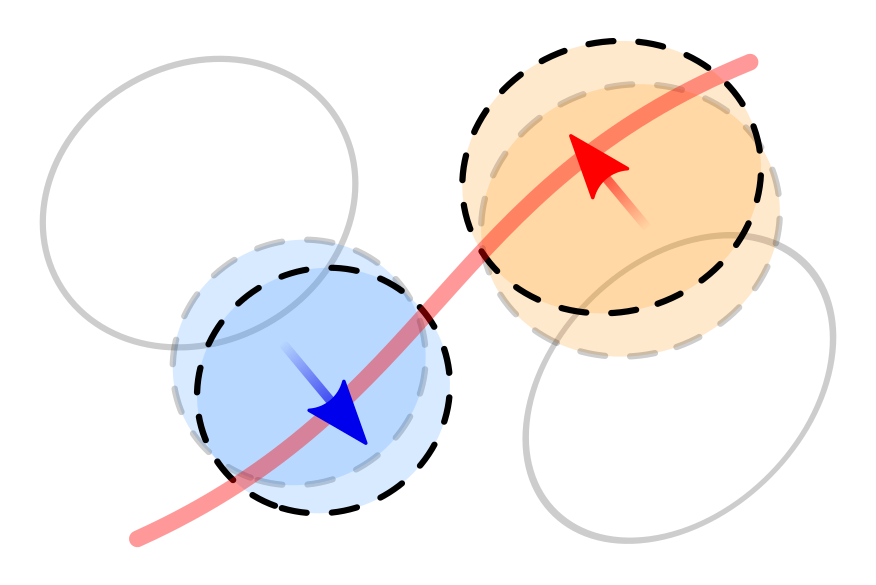}&
		\includegraphics[width=4cm]{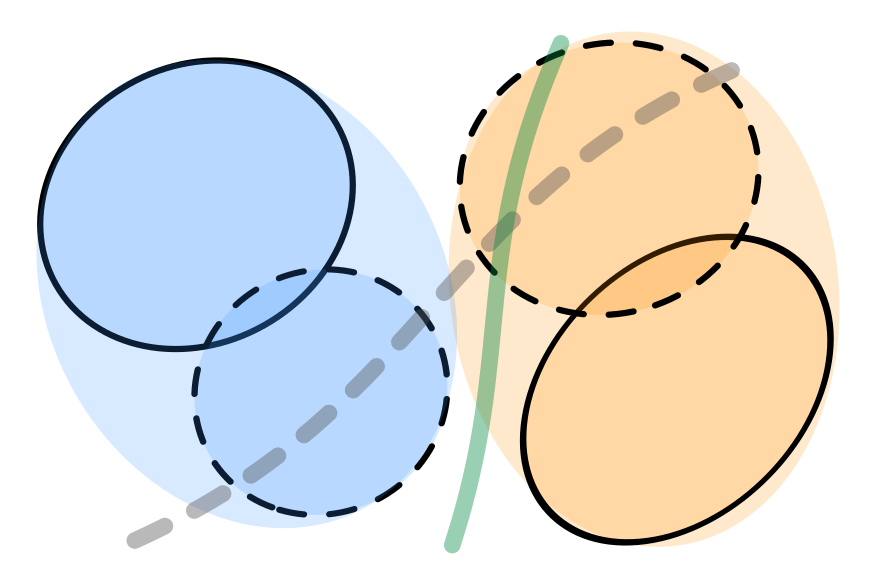}\\	
		\multicolumn{2}{c}{(c) AdD on feature extractor}	\vspace{0.2cm}\\
		\includegraphics[width=4cm]{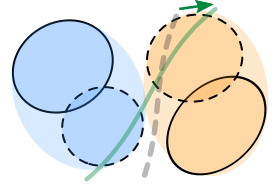}&
		\includegraphics[width=4cm]{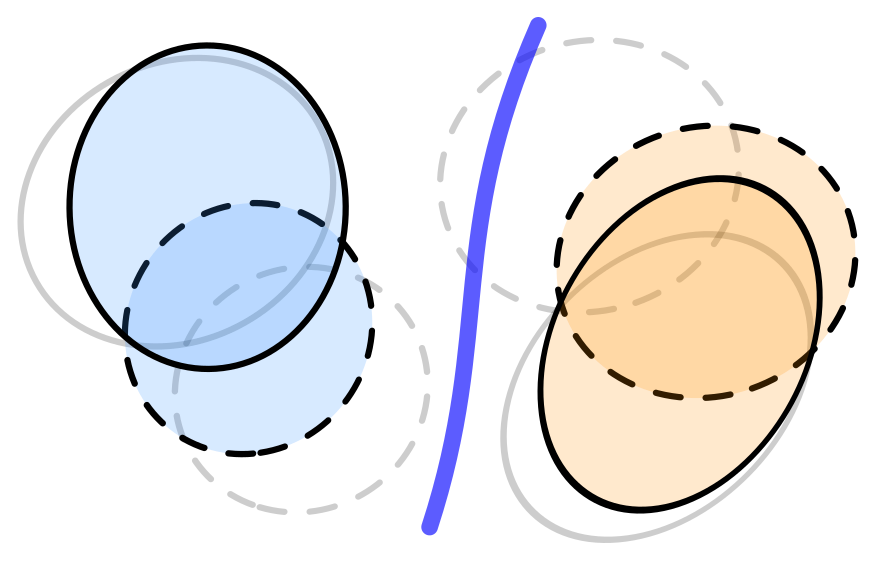}\\
		\multicolumn{2}{c}{(d) AdD on classifier}\\
	\end{tabular}\vspace{0.4cm}
	\caption{We illustrate the domain adaptation process with adversarial dropout (AdD). We depict the source and target domains as solid and dashed lines, respectively. {\color{red}Decision boundary} of a model only trained on the source domain easily violates the {\em cluster assumption} in that it passes through target feature-dense regions (a). We can apply AdD on both the feature extractor (c) and classifier (d). When AdD is used on the feature extractor, the decision boundary is pushed away from feature dense regions. On the contrary, AdD on the classifier pushes features away from the decision boundary. Eventually, our domain adapted model draws a {\color{blue}robust decision boundary} that avoids clusters (b).}\label{fig:overview}
\end{figure}

The advent of deep neural networks (DNNs) has shown exceptional performances on various visual recognition tasks using large-scale datasets~\cite{Deng2009ImageNet, Lin2014COCO, Geiger2013KITTI}. Training a DNN model begins with curating data and its associated label. In general, the annotation process is expensive and time-consuming. Moreover, we are unable to collect appropriate data in some cases, if events are rarely encountered or related to dangerous situations. Hence, researchers~\cite{visda2017, Richter2016GTA, VirtualKITTI2016, SynthiaData} are paying attention to leverage synthetic data in a simulation environment, where annotating labels is effortless to a wide range of scenarios. 

To take full advantage of synthetic datasets, domain adaptation has become an active research area. In the domain adaptation setting, we leverage rich annotations on a source domain to achieve strong performance on a target domain regardless of poor annotations. Nevertheless, a model trained only on the source domain provides disappointing outcomes when the target domain shows inherently different characteristics. 
This issue is known as \textit{domain shift} and is one of the main reasons for performance drops on the target domain. Therefore, we propose a novel method that can reduce the domain shift for domain adaptation.

In this paper, we tackle unsupervised domain adaptation (UDA), where the target domain is completely unlabelled. Recent works have proposed to align source and target domain distributions through domain adversarial training~\cite{Ganin2015, Tzeng2017, Ganin2016DANN}. These methods employ an auxiliary domain discriminator to obtain domain-invariant feature representation. 
The main assumption in domain adversarial training is that if the feature representation is domain-invariant, a classifier trained on the source domain's features will operate on the target domain as well. 
However, the weaknesses of domain adversarial methods have been pointed out in~\cite{Saito2018, Saito2018b, Shu2018}. Since the domain discriminator simply aligns source and target features without considering the class labels, it is likely that the resulting features will not only be domain-invariant, but also non-discriminative with respect to class labels. Consequently, it is hard to reach the optimal performance on classification.

Our approach is based on the cluster assumption, which states that decision boundaries should be placed in low density regions in the feature space~\cite{Chapelle2005}. Without model adaptation, the feature extractor generates indiscriminate features for unseen data from the target domain, and the classifier may draw decision boundaries that pass through feature-dense regions on the target domain. Thus, we learn a domain adapted model by pushing the decision boundary away from the target domain's features.  Our method, {\em Drop to Adapt} (DTA), employs adversarial dropout~\cite{Park2018} to enforce the cluster assumption on the target domain. More precisely, to support various tasks, we introduce element-wise and channel-wise adversarial dropout operations for fully-connected and convolutional layers, respectively. Fig.~\ref{fig:overview} overviews our method, and we design the associated loss functions in Section~\ref{ssec:train}. 

We summarize our contributions as follows: 1) We propose a generalized framework in UDA, which is built upon adversarial dropout~\cite{Park2018}. Our implementation supports both convolutional and fully connected layers; 2) We test on various domain adaptation benchmarks for image classification, and achieve competitive results compared to state-of-the-art methods; and 3) We extend the proposed method to a semantic segmentation task in UDA, where we perform adaptation from the simulation to real-world environments.

\section{Related Work}\label{sec:realted}
\paragraph{Domain adaptation} has been studied extensively. Ben-David \textit{et al.}~\cite{Ben2010, Ben2006} examined various divergence metrics between two domains, and defined an upper bound for the target domain error. Based on these studies, image-translation methods minimize the discrepancy between the two domains at an image-level \cite{Taigman2017DTN, Zhu2017, Bousmalis2017}.

On the other hand, feature alignment methods have attempted to match feature distributions between the source and target domains\cite{Ganin2015, Tzeng2017, Long2015DAN}. In particular, Ganin \textit{et al.}~\cite{Ganin2015} proposed a domain adversarial training method that aims to generate domain-invariant features by deceiving a domain discriminator. Many recent works use domain adversarial training as a key component in their adaptation procedure \cite{Ganin2016DANN, Bousmalis2016, Hoffman2018, Shu2018, Pei2018MADA, Wang2019TADA, Volpi2018AFA}. However, the domain classifier cannot consider class labels; thus, the generated features tend to be sub-optimal for classification.

To overcome the weaknesses of domain adversarial training, more recent works directly deal with the relationship between the decision boundary and feature representations based on the cluster assumption~\cite{Chapelle2005}. Several works~\cite{Long2016RTN, French2018, Shu2018} exploit semi-supervised learning for domain adaptation. Besides, MCD~\cite{Saito2018} and ADR~\cite{Saito2018b} use a minimax training method to push target feature distributions away from the decision boundary, where both methods are composed of the feature extractor and the classifiers. More precisely, in~\cite{Saito2018b}, two different classifiers are sampled via stochastic dropout. Then, for the same target data sample, the classifiers are updated to maximize the discrepancy between the two predictions. Lastly, the feature extractor is updated multiple times to minimize this discrepancy. The minimax training process leaves the classifier in a noise sensitive state. Therefore, it must be newly trained for optimal performance. 

Though our work is partly inspired by ADR, the proposed method is more efficient and simpler to train compared to the prior arts~\cite{Saito2018b,Saito2018}. Instead of updating the classifier for maximizing discrepancy, we employ adversarial dropout~\cite{Park2018} on the classifier to achieve a similar effect. Furthermore, this adversarial dropout can be applied to the feature extractor as well. Without the need of a minimax training scheme, DTA has a straightforward and reliable adaptation process.

\vspace{-0.3cm}
\paragraph{Dropout} is a simple yet effective regularization method that randomly drops a fraction of the neurons during the training process~\cite{Srivastava2014Dropout}. According to Srivastava \textit{el al.}~\cite{Srivastava2014Dropout}, dropout has the effect of ensembling multiple subsets of a network.  Park~\etal~\cite{Park2014Dropout} spotlighted the efficacy of the dropout on convolutional layers.
Tompson \textit{el al.}~\cite{Tompson2015SpatialDropout} pointed out that activations of convolutional layers are usually surrounded by similar activations within the same feature map; thus, dropping individual neurons does not have a strong effect in convolution layers. Instead, they proposed spatial dropout, which drops entire feature maps instead of individual neurons. Building on spatial dropout, Hou \textit{el al.}~\cite{Hou2019WCD} proposed a weighted channel dropout that uses variable drop rates for individual channels, where the drop rates depend on the channel's averaged activation value. The weighted channel dropout is only applied to deep layers of the network, where activations are known to have high specificity \cite{Zhang2016FineGrained, Zeiler2014VisCNN, Yosinski2014TransfeFea}. Similarly, for channel-wise adversarial dropout, we remove entire feature maps in an adversarial way.

\section{Proposed Method}\label{sec:algo}

\subsection{Unsupervised Domain Adaptation}\label{ssec:uda}

We first define the unsupervised domain adaptation (UDA) problem in general, and relevant notations to our work. In the UDA setting, we use data from two distinctive domains: the source domain $\sourcedomain = \{X_s, Y_s\}$ and the target domain $\targetdomain = \{X_t\}$. A data point from the source domain $\bx{s}\in X_s$ has an associated label $y_s\in Y_s$,  whereas one from the target domain $\bx{t}\in X_t$ has no paired ground-truth label. We employ a feature extractor $f(\bx{}; \dropmask_f)$, where $\dropmask_f$ represents a dropout mask which can be applied at an arbitrary layer of the feature extractor. The feature extractor takes a data point from two domains $\bx{}\sim\sourcedomain\cup\targetdomain$ and creates a latent vector, which is fed into a classifier $c(\cdot;\dropmask_c)$. The classifier applies a dropout mask $\dropmask_c$ at an arbitrary layer. We denote the entire neural network as a composition of the feature extractor and the classifier: $h(\bx{}; \dropmask_f, \dropmask_c) = c(f(\bx{}; \dropmask_f); \dropmask_c)$.

\subsection{Adversarial Dropout}\label{ssec:dta}
We leverage a non-stochastic dropout mechanism, Adversarial Dropout (AdD)~\cite{Park2018}, for unsupervised domain adaptation. Adversarial dropout was originally proposed as an effective regularization method for supervised and semi-supervised learning. More specifically, Park \etal~\cite{Park2018} define two types of Adversarial Dropout: Supervised Adversarial Dropout (SAdD), and Virtual Adversarial Dropout (VAdD). With access to ground truth labels, SAdD is used to maximize the divergence between a model's prediction and ground truth label. Without labels, on the other hand, VAdD is used to maximize the divergence between two independent predictions to an input. Due to the lack of target domain labels, SAdD cannot be employed for our purpose. Thus, we exclusively work with VAdD, which is referred to as AdD for the sake of convenience.

AdD provides a simple and efficient mechanism of generating two divergent predictions for an input. Ultimately, our goal is to enforce the cluster assumption on target data by minimizing the divergence between predictions. To this end, we introduce element-wise AdD (EAdD) and propose its variant, channel-wise AdD (CAdD). 

We first define a dropout mask $\dropmask$ applied to an intermediate layer of a network $h$. For simplicity, we decompose a network $h$ into the subsequent sub-networks $\lowernet$ and $\uppernet$ by the layer applied dropout $\dropmask$, such as:
\begin{equation}
	h(\bx{}; \dropmask) = \uppernet(\dropmask \odot \lowernet(\bx{})),
\end{equation}
where $\odot$ represents the element-wise multiplication. Note that $\dropmask$ has the same dimensions to the output of $\lowernet(\bx{})$. 

Let $D[p,p']\geq 0$ measure the divergence between two distributions $p$ and $p'$. Then, the divergence between the predictions of $\bx{}$ with different dropout masks, $\dropmask$ and $\dropmask^s$, is defined as:
\begin{align}
	D&\left[h(\bx{};\dropmask^s),h(\bx{};\dropmask)\right]\\\nonumber
	&=D\left[\uppernet(\dropmask^s\odot\lowernet(\bx{})), \uppernet(\dropmask\odot\lowernet(\bx{}))\right].
\end{align}

\subsubsection{Element-wise Adversarial Dropout}
The element-wise adversarial dropout (EAdD) mask $\dropmask^{adv}$ is defined with respect to a stochastic dropout mask $\dropmask^{s}$ as:
\begin{align}
	&\dropmask^{adv} = \argmax_{\dropmask}D\left[h(\bx{};\dropmask^s), h(\bx{};\dropmask)\right]\nonumber\\
	&\text{where $\|{\dropmask^s}-\dropmask\|\leq\delta_e L$,}
\end{align}
where $L$ denotes the dimension of $\dropmask\in\mathbb{R}^L$, and $\delta_e$ is a hyper parameter to control the perturbation magnitude with respect to $\dropmask^s$. The objective is to find a minimally modified adversarial mask $\dropmask^{adv}$ that maximizes the output divergence $D$ between two independent forward passes of $\bx{}$. 

To find $\dropmask^{adv}$, Park \etal~\cite{Park2018} optimize a 0/1 knapsack problem with appropriate relaxations in the process. Their optimization process can be simplified into the following steps. First, an \textit{impact value} is approximated for each element in $\lowernet(\bx{})$, which is directly proportional to the element's contribution for increasing the divergence. When negative, the element has a decreasing effect on the divergence. Then, without breaching the boundary condition, the elements of $\dropmask^s$ are adjusted to maximize divergence. 

\begin{figure}[t]
	\centering
	\subfigure[Element-wise AdD (EAdD)]{\includegraphics[width=4cm]{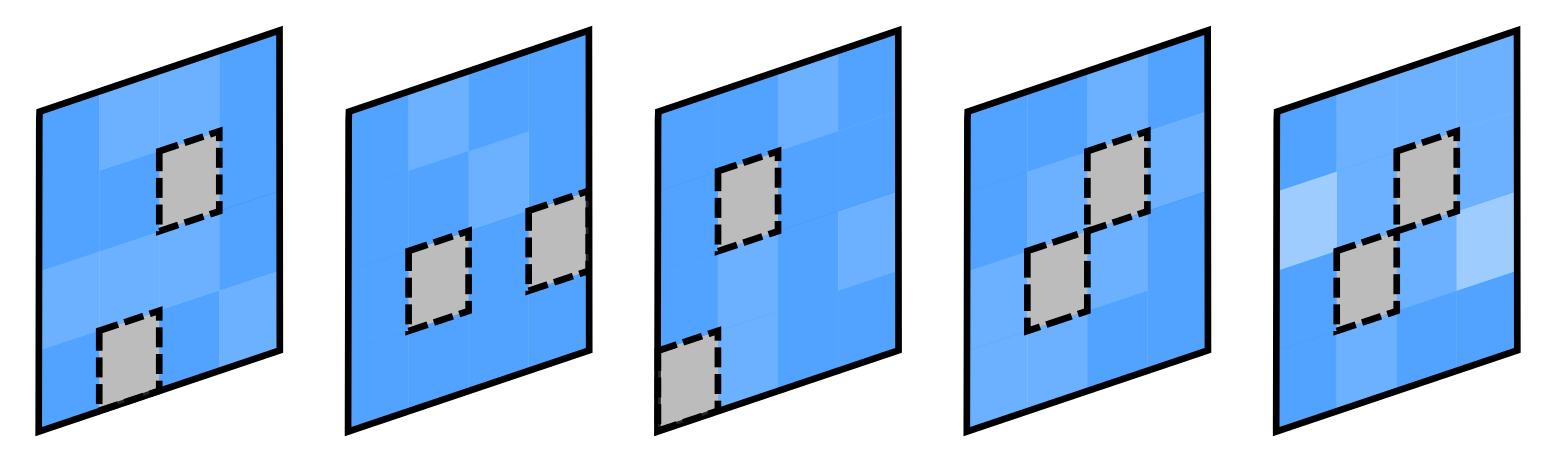}}
	\subfigure[Channel-wise AdD (CAdD)]{\includegraphics[width=4cm]{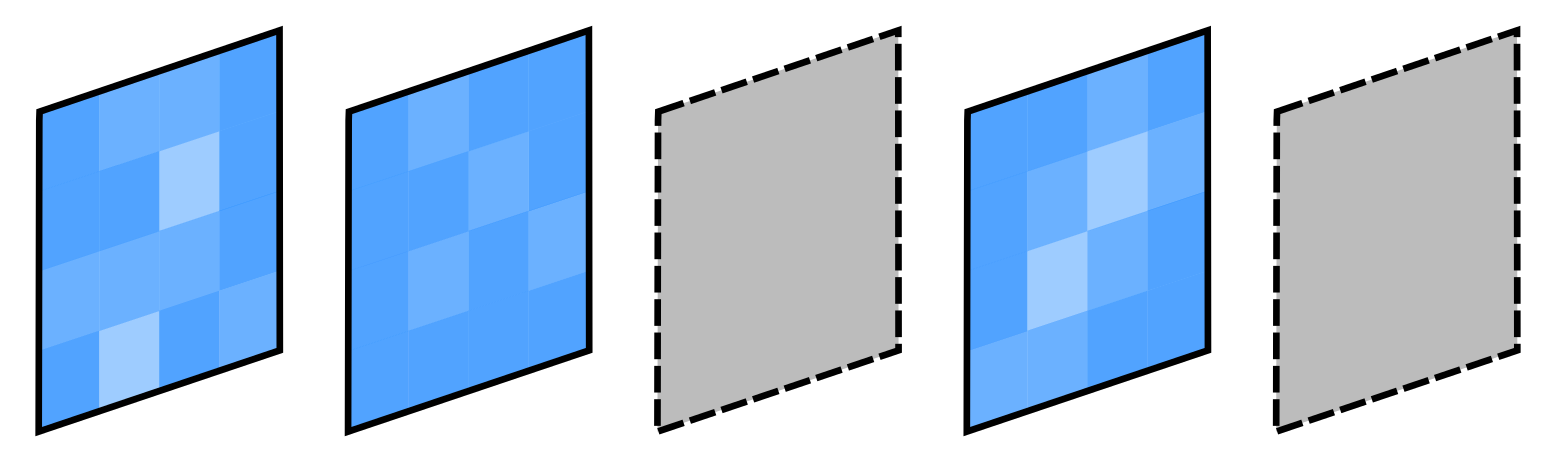}}	
	\caption{\textbf{Comparison of EAdD and CAdD.} EAdD drops units individually, regardless of spatial correlation. CAdD, on the other hand, drops entire feature maps, making it more suitable for convolutional layers.}
	\label{fig:dropout}
\end{figure}

\subsubsection{Channel-wise Adversarial Dropout}
To use \methodname~in a wider range of tasks, we extend EAdD to convolutional layers. In these layers, however, standard dropout is relatively ineffective due to the strong spatial correlation between individual activations of a feature map~\cite{Tompson2015SpatialDropout}. EAdD dropout suffers from the same issues when naively applied to convolutional layers. 

Hence, we formulate CAdD, which adversarially drops entire feature maps rather than individual activations. While the general procedure is similar to that of EAdD, we impose certain constraints on the mask to represent spatial dropout~\cite{Tompson2015SpatialDropout}. Fig.~\ref{fig:dropout} highlights the difference between EAdD and CAdD.

Consider the activation of a convolutional layer, $\lowernet(\bx{}) \in \mathbb{R}^{C \times H \times W}$, where $C$, $H$, and $W$ denote the channel, height, and width dimensions of the activation, respectively. We define a channel-wise dropout mask $\dropmask(i) \in \mathbb{R}^{ H \times W}$, with the following constraints:
\begin{align}
	\dropmask(i) = \boldsymbol{0}\ \text{or}\ \boldsymbol{1},\forall i \in\{1,\cdots,C\}.\label{mask_contraints}
\end{align}
Here, $\dropmask(i)$ corresponds to the $i$-th activation map of $\lowernet(\bx{})$, $\boldsymbol{0}\in\mathbb{R}^{H \times W}$ denotes a matrix of zeros, and $\boldsymbol{1}\in\mathbb{R}^{H \times W}$ denotes a matrix of ones, respectively. Then, the channel-wise adversarial dropout mask is defined as:
\begin{align}
	&\dropmask^{adv} = \argmax_{\dropmask}D\left[h(\bx{};\dropmask^s), h(\bx{};\dropmask)\right],\nonumber\\
	&\text{where $\frac{1}{HW}\sum_{i=1}^{C}{\|{\dropmask^s(i)}-\dropmask(i)\|} \leq \delta_c C$}.
\end{align}
As before, $\delta_c$ is the hyper parameter that controls degree of the perturbation. 

The process of finding the channel-wise adversarial dropout mask $\dropmask^{adv}$ is similar to those of element-wise adversarial dropout. For CAdD, however, the impact value is approximated for each activation map of $\lowernet(\bx{})$ due to the constraints in Eq.~\eqref{mask_contraints}. We provide the further details about the approximation in Appendix A of our supplementary material.

\subsection{Drop to Adapt}\label{ssec:train}
Unlike the prior arts~\cite{Saito2018,Saito2018b}, the proposed algorithm leverages a unified objective function to optimize all network parameters. The overall loss function is defined as a weighted sum of four objective functions:
\begin{align}
\lossfunc{}(\sourcedomain, \targetdomain) 
= \lossfunc{T}(\sourcedomain) 
+ \lambda_1 \lossfunc{DTA}(\targetdomain)
+ \lambda_2 \lossfunc{E}(\targetdomain)
+ \lambda_3 \lossfunc{V}(\targetdomain)
,
\end{align}
where $\lossfunc{T}$, $\lossfunc{DTA}$, $\lossfunc{E}$, and $\lossfunc{V}$ represent the objectives for task-specific, domain adaptation, entropy minimization and Virtual Adversarial Training (VAT)~\cite{Miyato2018}, respectively. Also, the associated hyper-parameters, $\lambda_1$, $\lambda_2$, and $\lambda_3$, control the relative importance of the terms. 

\paragraph{Task-specific objective.} We define the task-specific objective function $\lossfunc{T}$ regarding the source domain $\sourcedomain$. In practice, this objective function can be replaced according to the given task. As an example, we present the cross entropy which is widely used for classification:
\begin{equation}
\lossfunc{T}(\sourcedomain) = -\mathbb{E}_{\bx{s}, y_s\sim \sourcedomain}[\by{s}^T\log{h(\bx{s})}],
\end{equation}
where $\by{s}$ is one-hot encoded vector of $y_s$.

\paragraph{Domain adaptation objective.} 
As the main component, we present the objective function for the domain adaptation first. The objective consists of two parts to affect on the feature extractor $\lossfunc{fDTA}$ and the classifier $\lossfunc{cDTA}$:
\begin{equation}
\lossfunc{DTA}(\targetdomain) = \lossfunc{fDTA}(\targetdomain) + \lossfunc{cDTA}(\targetdomain).
\end{equation}

We aim to minimize the divergence between two predicted distribution regarding to an input $\bx{}$: one with a random dropout mask $\dropmask^s_f$ and another with an adversarial dropout mask $\dropmask^{adv}_f$. Among the various divergence measures, we choose the Kullback-Leibler (KL) divergence in this work. Assuming that the feature extractor consists of convolutional layers, we employ channel-wise adversarial dropout for $\dropmask^{adv}_f$:
\begin{align}
\lossfunc{fDTA}(\targetdomain)
&= \mathbb{E}_{\bx{t}\sim \targetdomain}\Big[D\left[h(\bx{t}; \dropmask^s_f),h(\bx{t}; \dropmask^{adv}_f)\right]\Big]\nonumber\\
&= \mathbb{E}_{\bx{t}\sim \targetdomain}\Big[D_{KL}\left[h(\bx{t};\dropmask^s_f)\|h(\bx{t}; \dropmask^{adv}_f))\right]\Big].
\end{align}

We illustrate the effects of $\lossfunc{fDTA}$ in Fig.~\ref{fig:overview}(c). Initially, the decision boundary crosses high density regions in the feature space (Fig.~\ref{fig:overview}(a)), which is in violation of the cluster assumption. By applying adversarial dropout on the feature extractor, we cause certain features to cross the decision boundary (Fig.~\ref{fig:overview}(c), left). Then, to enforce consistent predictions, the model parameter are updated to push the decision boundary away from these features (Fig.~\ref{fig:overview}(c), right). 

Similarly, we apply AdD to the classifier, where the classifier is defined as a series of fully connected layers. Thus, we perform the element-wise adversarial dropout $\dropmask^{adv}_c$ and compute the divergence:
\begin{align}
\lossfunc{cDTA}(\targetdomain) = \mathbb{E}_{\bx{t}\sim \targetdomain}\Big[D_{KL}\left[h(\bx{t};\dropmask^s_c)\|h(\bx{t}; \dropmask^{adv}_c))\right]\Big].
\end{align}

When adversarial dropout is applied on the classifier, we determine the most volatile areas in the feature space. These volatile regions are in the vicinity of the decision boundary, and predictions in these regions can be changed even by a small perturbation. (Fig.~\ref{fig:overview}(d), left). Therefore, minimizing $\lossfunc{cDTA}$ lets the features avoid falling into such volatile regions (Fig.~\ref{fig:overview}(d), right). 


\paragraph{Entropy minimization objective.}
We introduce the entropy minimization objective to enforce the cluster assumption further. This loss penalizes target samples for being close to the decision boundary, and thus, causes the model to learn more discriminative features:
\begin{equation}
\lossfunc{E}(\targetdomain) = - \mathbb{E}_{\bx{t}\sim \targetdomain} [h(\bx{t})^T \log{h(\bx{t})}].\label{eq:ent}
\end{equation}

\paragraph{VAT objective.}
Lastly, we exploit VAT, which adversarially perturbs the target data at the input level. The VAT minimization objective is defined as:
\begin{equation}
\lossfunc{V}(\targetdomain) = \mathbb{E}_{\bx{t}\sim \targetdomain} \Big[\max_{\|r\| \leq \epsilon} D_{KL}\left[h(\bx{t}) \| h(\bx{t} + r)\right]\Big],
\end{equation}
where $r$ represents the virtual adversarial perturbation on input $\bx{t}$. While DTA and VAT are similarly motivated, they regularize the network with different forms of perturbations: network parameter perturbations (DTA) and input perturbations (VAT). Thus, VAT provides an orthogonal regularization to DTA, leading to complementary effects.

\begin{figure}[h]
	\centering
	\footnotesize
	\begin{tabular}{c@{\hskip1pt}c@{\hskip1pt}c@{\hskip1pt}c@{\hskip-.5pt}c}
		\includegraphics[width=1.55cm]{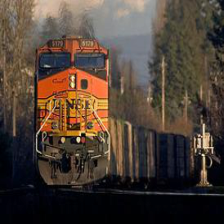}&%
		\includegraphics[width=1.55cm]{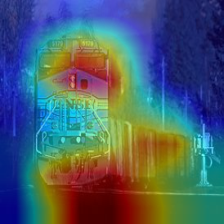}&%
		\includegraphics[width=1.55cm]{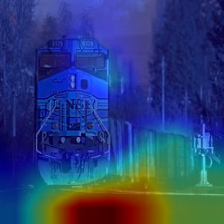}&%
		\includegraphics[width=1.55cm]{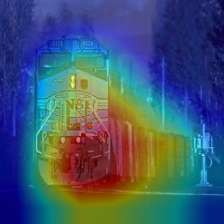}&%
		\includegraphics[width=1.55cm]{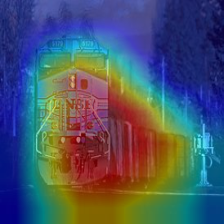}\\[-0.2em]
		\includegraphics[width=1.55cm]{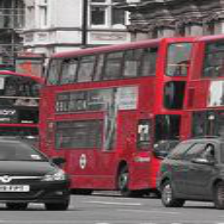}&%
		\includegraphics[width=1.55cm]{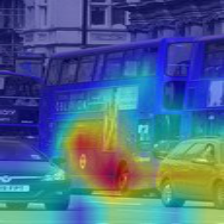}&%
		\includegraphics[width=1.55cm]{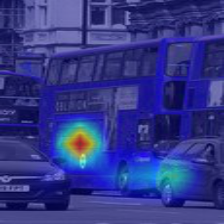}&%
		\includegraphics[width=1.55cm]{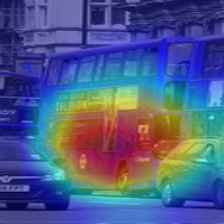}&%
		\includegraphics[width=1.55cm]{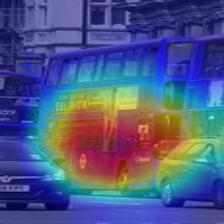}\\[-0.2em]
		\includegraphics[width=1.55cm]{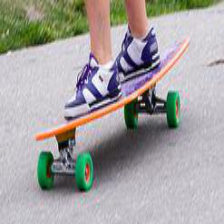}&%
		\includegraphics[width=1.55cm]{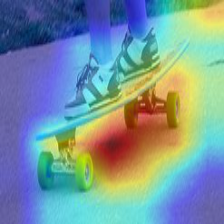}&%
		\includegraphics[width=1.55cm]{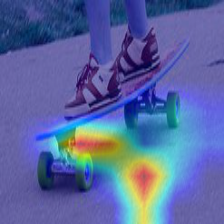}&%
		\includegraphics[width=1.55cm]{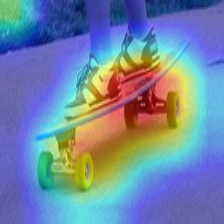}&%
		\includegraphics[width=1.55cm]{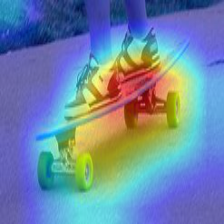}\\[-0.2em]
		(a) Input & (b) SO & (c) SO+AdD  & (d) DTA & (e) DTA+AdD\\
	\end{tabular}\vspace{0.2cm}
	\caption{{\bf Effect of adversarial dropout.} We visualize class activation maps on target domain images using GradCAM~\cite{Selvaraju_2017_ICCV}. Adversarial dropout (c) effectively deactivates semantically meaningful regions for a prediction compared to its baseline model only trained on source domain (b). Our domain adapted model (DTA) produces reasonable predictions (d), even though $10\%$ of units are eliminated by AdD (e).}\label{fig:gradcam}
\end{figure}

\paragraph{Interpretation of DTA.}
Fig.~\ref{fig:gradcam} visualizes the effects of adversarial dropout using Grad-GAM~\cite{Selvaraju_2017_ICCV}, which accentuates the most discriminative regions for a prediction. As a baseline, we present Grad-CAM visualizations of a model trained only on the source domain (SO, see Fig.~\ref{fig:gradcam}(b)). We apply AdD on the source only model (SO + AdD), and see that semantically meaningful areas are deactivated. In contrast, our domain adapted model (DTA, see Fig.~\ref{fig:gradcam}(d)) stays relatively unaffected by AdD, as it keeps seeing the same discriminative regions (see Fig.~\ref{fig:gradcam}(e)) regardless of AdD. The visualizations imply that AdD promotes activations on more hidden units, and lends to robust decision boundary across the domains.


\section{Experimental Results}\label{sec:results}
\definecolor{lightgray}{gray}{0.9}
\begin{table}[t]
	\caption
	{
		Results of experiment on small image datasets. *We compare with the MT+CT+TF for SE.
	}
	\vspace{0.2cm}
	\centering
	\resizebox{0.48\textwidth}{!}{
		\begin{tabular}{l c c c c c}	
			\hline\hline
			Source & SVHN & MNIST & USPS  & STL & CIFAR \\
			Target & MNIST & USPS & MNIST  & CIFAR &  STL \\
			\hline\hline
			Source only (Ours)  & 76.5 & 96.3    & 76.9  & 60.1 & 78.2\\
			\hline
			SE*~\cite{French2018}& 98.6 & 98.1    & 97.3   & 74.2  & 79.7 \\
			VADA~\cite{Shu2018}& 94.5 & -       & -           & 73.5 & 80.0 \\
			DIRT-T~\cite{Shu2018}& 99.4& -       & -        & 75.5 & -       \\
			Co-DA~\cite{CorregulNIPS2018}& 98.3 & -   & -  & \textbf{76.4}  & 81.1   \\
			Co-DA+DIRT-T~\cite{CorregulNIPS2018} & 99.4  & - & -  & 76.3  & - \\ 
			Ours                & \textbf{99.4}& \textbf{99.5}    & \textbf{99.1}    & 72.8 & \textbf{82.6}    \\
			\hline		
			Target only (Ours)  & 99.6& 97.8    & 99.6   & 90.4 & 70.0  \\ 
			\hline\hline
	\end{tabular}}
	\label{tab:small_image}
\end{table}

In this section, we evaluate the proposed method on small and large DA benchmarks. To demonstrate the generality of our model, we conduct the experiments in two major recognition tasks: classification and segmentation. In each experiment, we select one domain as the source domain, and another as the target domain. We denote "Source only" as the target domain performance of a model trained on the source domain, and "Target only" as that of a model trained on the target domain. These two serve as baselines for the lower and upper bound performance in domain adaptation. We do not tune a set of data augmentation schemes nor do we report performance with ensemble predictions, as in French \textit{el al.}~\cite{French2018}. Rather, all evaluation results are based on the same data augmentation strategy with a single model prediction.


\begin{table*}[t]
	\caption
	{
		Results on VisDA-2017 classification using ResNet-101.
	}
	\vspace{0.2cm}
	\centering
	\resizebox{0.85\textwidth}{!}{
		\begin{tabular}{l | c c c c c c c c c c c c | c }	
			\hline\hline
			& aero. & bike & bus & car & horse & knife & moto. & person & plant & sktb. & train & truck & avg.\\
			\hline\hline
			Source Only & 46.2 & 27.6 & 31.4 & 78.1 & 71.8 & 1.3 & 71.7 & 14.3 & 63.5 & 31.0 & 93.7 & 3.2 & 50.8 \\
			\hline
			DAN~\cite{Long2015DAN} & 87.1 & 63.0 & 76.5 & 42.0 & 90.3 & 42.9 & 85.9 & 53.1 & 49.7 & 36.3 & 85.8 & 20.7 & 61.1 \\
			DANN~\cite{Ganin2015} & 81.9 & 77.7 & 82.8 & 44.3 & 81.2 & 29.5 & 65.1 & 28.6 & 51.9 & 54.6 & 82.8 & 7.8 & 57.4 \\
			MCD~\cite{Saito2018} & 87.0 & 60.9 & 83.7 & 64.0 & 88.9 & 79.6 & 84.7 & 76.9 & 88.6 & 40.3 & 83.0 & 25.8 & 71.9 \\
			ADR~\cite{Saito2018b} & 87.8 & 79.5 & 83.7 & 65.3 & 92.3 & 61.8 & 88.9 & 73.2 & 87.8 & 60.0 & 85.5 & \textbf{32.3} & 74.8 \\
			\hline\hline
			Ours & \textbf{93.7} & \textbf{82.2} & \textbf{85.6} & \textbf{83.8} & \textbf{93.0} & \textbf{81.0} & \textbf{90.7} & \textbf{82.1} & \textbf{95.1} & \textbf{78.1} & \textbf{86.4} & 32.1 & \textbf{81.5} \\
			\hline\hline
	\end{tabular}}
	\label{tab:visda_results}
\end{table*}

\subsection{DA on Small Datasets}\label{ssec:small_image}

To evaluate the influence of \methodname\ model, we first perform experiments on small datasets. We use MNIST~\cite{Lecun1998}, USPS~\cite{Hull1994}, and Street View House Numbers (SVHN)~\cite{Netzer2011} for adaptation on digits recognition. For object recognition, we use CIFAR10 (CIFAR)~\cite{Krizhevsky2009} and STL10 (STL)~\cite{Coates2011}. For fair comparison against recent state-of-the-art methods such as Self-Ensembling (SE)~\cite{French2018}, VADA~\cite{Shu2018}, and DIRT-T~\cite{Shu2018}, we conduct experiments on the same network architecture as in SE. Not that while VADA/DIRT-T use a slightly differernet architecture, the total number of parameters are comparable. The results can be found in Table~\ref{tab:small_image}, and a full list of hyperparameter settings can be found in Appendix B.

\paragraph{SVHN $\rightarrow$ MNIST.}
SVHN and MNIST are two digit classification datasets with a drastic distributional shift between the two. While MNIST consists of binary handwritten digit images, SVHN consists of colored images of street house numbers. Since MNIST has a significantly lower image dimensionality than SVHN, we adopt the dimension of MNIST to 32 $\times$ 32 of SVHN, with three channels. When the proposed \methodname\ is applied, our approach demonstrates a significant improvement over previous works, and achieves a performance similar to the "Target only" performance on MNIST. 

\paragraph{MNIST $\leftrightarrow$ USPS.}
MNIST and USPS contain grayscale images, so the domain shift between these two datasets is relatively smaller compared to that of the SVHN $\rightarrow$ MNIST setting. In both adaptation directions, we achieve an accuracy close to the performance of fully supervised learning on the target domain. In fact, we obtain higher accuracy on USPS when adapting from MNIST, than when trained directly on USPS. This is because the USPS training is relatively small, allowing us to achieve improved performance by adapting from MNIST, using \methodname. 

\paragraph{CIFAR $\leftrightarrow$ STL.}
CIFAR and STL are 10-class object recognition datasets with colored images. We remove the non-overlapping classes and redefine the task as a 9-class classification task. Furthermore, we downscale the 96 $\times$ 96 image dimesion of STL to match the 32 $\times$ 32 dimension of CIFAR. In the \textbf{CIFAR $\rightarrow$ STL} setting, our method's performance surpasses others by a comfortable margin. For the same reasons presented in the \textbf{MNIST $\rightarrow$ USPS} setting, our adapted model outperforms the target only model on this dataset pair. In \textbf{STL $\rightarrow$ CIFAR}, however, our method is slightly weak. This is because STL contains a very small dataset, with only 50 images per class. Since \methodname\ regularizes the decision boundary of the model, the inherent assumption is that the model can achieve low generalization error on the source domain. This assumption holds in most cases, but breaks down when STL is the source domain.

To summarize, we achieve a substantial margin of improvement over the source only model across all domain configurations. In four of the five configurations, our method outperforms the recent state-of-the-art results. Next, we evaluate our method on more practical settings that embody real-life domain adaptation scenarios.

\begin{table*}[t]
	\caption
	{
		Results on GTA $\rightarrow$ Cityscapes, using a modified FCN with ResNet-50 as the base network.
	}
	\vspace{0.2cm}
	\centering
	\resizebox{1.0\textwidth}{!}{
		\begin{tabular}{l | c c c c c c c c c c c c c c c c c c c | c}	
			\hline\hline
			& \rotatebox{90}{road} & \rotatebox{90}{sidewalk} & \rotatebox{90}{building} & \rotatebox{90}{wall} & \rotatebox{90}{fence} & \rotatebox{90}{pole} & \rotatebox{90}{t light} & \rotatebox{90}{t sign} & \rotatebox{90}{veg} & \rotatebox{90}{terrian} & \rotatebox{90}{sky} & \rotatebox{90}{person} & \rotatebox{90}{rider} & \rotatebox{90}{car} & \rotatebox{90}{truck} & \rotatebox{90}{bus} & \rotatebox{90}{train} & \rotatebox{90}{mbike} & \rotatebox{90}{bike} & \rotatebox{90}{mIoU}\\
			\hline\hline
			Source only  &25.3 & 13.7 & 56.8 & 2.7 & \textbf{17.2} & \textbf{21.2} & 20.0 & \textbf{8.7} & 75.3 & 11.2 & 72.0 & 45.7 & 4.9 & 42.2 & 14.2 & 20.2 & 0.4 & \textbf{19.5} & 0.0 & 24.8\\
			DANN  & 72.4 & 19.1 & 73.0 & 3.9 & 9.3 & 17.3 & 13.1 & 5.5 & 71.0 & 20.1 & 62.2 & 32.6 & 5.2 & 68.4 & 12.1 & 9.9 & 0.0 & 5.8 & 0.0 & 26.4\\
			ADR  & 87.8 & 15.6 & \textbf{77.4} & 20.6 & 9.7 & 19.0 & 19.9 & 7.7 & 82.0 & \textbf{31.5} & 74.3 & 43.5 & 9.0 & \textbf{77.8} & 17.5 & 27.7 & \textbf{1.8} & 9.7 & 0.0 & 33.3\\
			\hline\hline
			Ours  & \textbf{88.8} & \textbf{36.9} & 76.9 & \textbf{20.9} & 15.4 & 19.6 & \textbf{21.8} & 7.9 & \textbf{82.9} & 26.7 & \textbf{76.1} & \textbf{51.7} & \textbf{9.4} & 76.1 & \textbf{22.4} & \textbf{28.9} & 1.7 & 15.2 & 0.0 & \textbf{35.8}\\
			\hline\hline
	\end{tabular}}
	\label{tab:semantic_results}
\end{table*}

\begin{figure}[t]
	\centering
	\footnotesize
	\begin{tabular}{c@{\hskip1pt}c}
		\includegraphics[width=4cm]{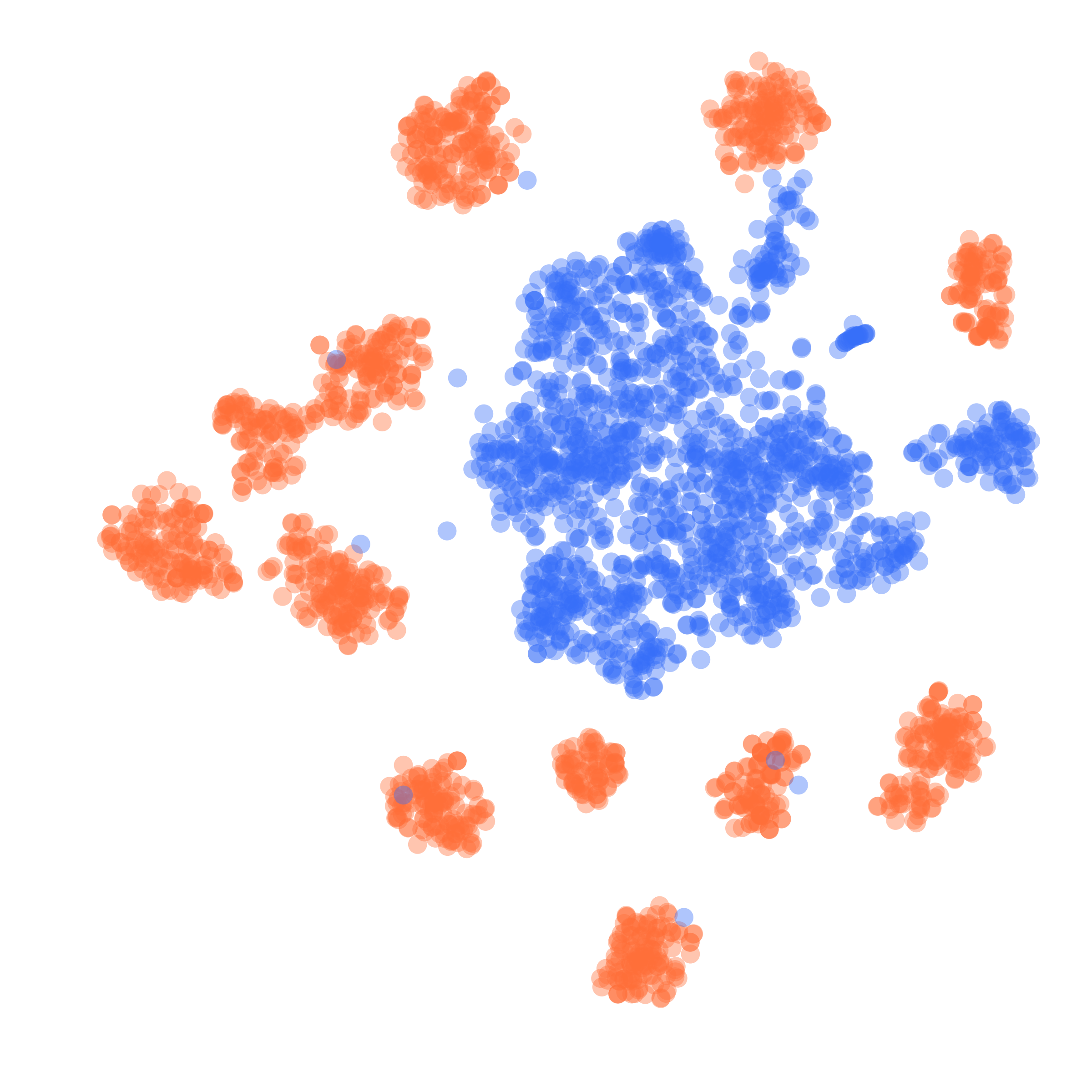}&%
		\includegraphics[width=4cm]{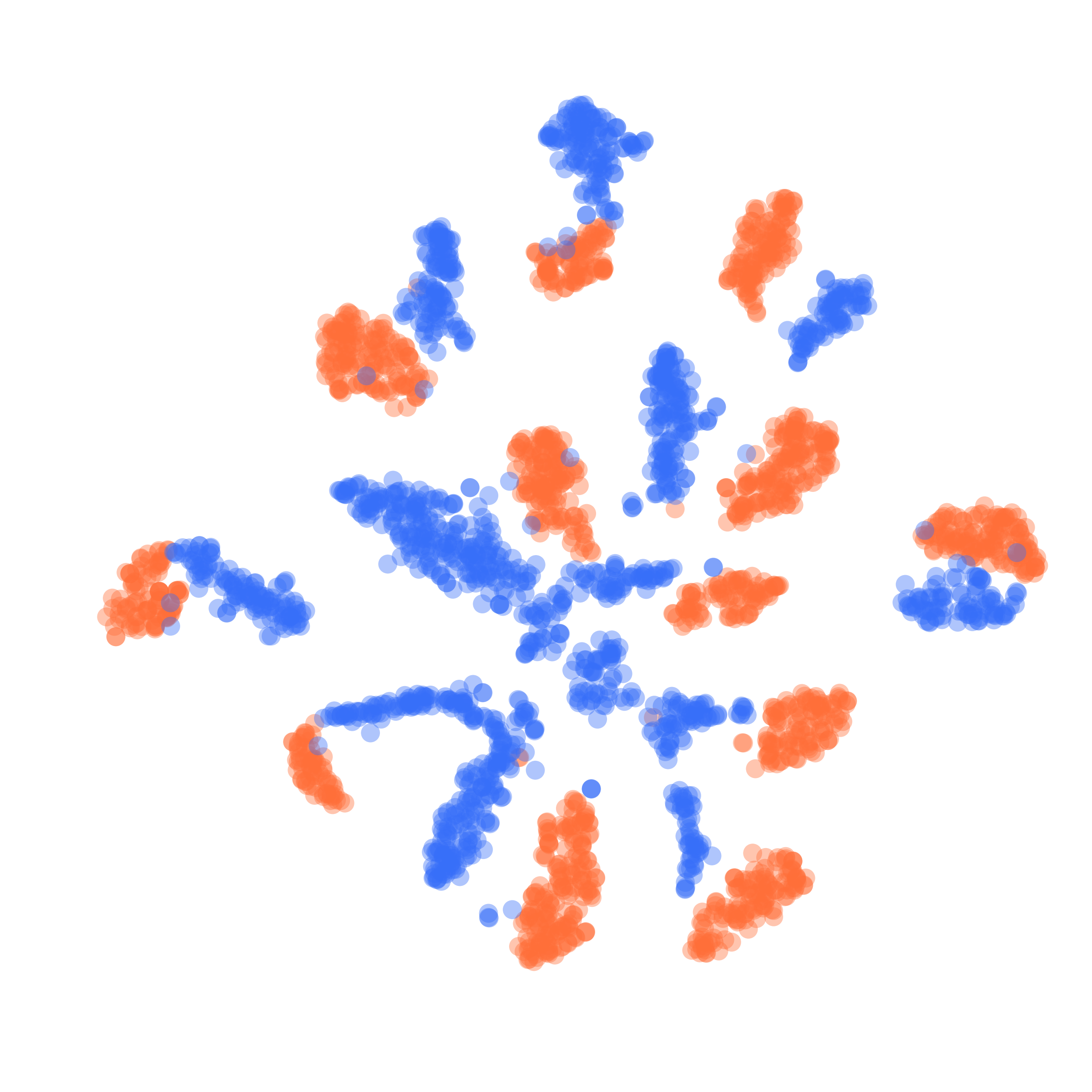}\\
		(a) Source Only & (b) \methodname \\
	\end{tabular}\vspace{0.3cm}
	\caption{{\bf t-SNE.} t-SNE visualization of VisDA-2017 classification dataset using ResNet-101, before and after adaptation with \methodname. t-SNE hyperparameters are consistent in both visualizations.}\label{fig:tsne}
\end{figure}

\begin{table}[t]
	\caption
	{
		Results on VisDA-2017 classification using ResNet-50. *SE report ensemble of multiple predictions. All other methods, including ours, report the average achieved by a single prediction. 
	}
	\vspace{0.2cm}
	\centering
    \scalebox{0.85}{%
	\begin{tabular}{l | c }	
		\hline\hline
		Method & avg.\\
		\hline\hline
		Source Only (Ours) & 45.6 \\
		\hline
		DAN~\cite{Long2015DAN} & 53.0 \\
		RTN~\cite{Long2016RTN} & 53.6 \\
		DANN~\cite{Ganin2015} & 55.0 \\
		JAN-A~\cite{Long2017JAN} & 61.6 \\
		GTA~\cite{Sank2018GTA} & 69.5 \\
		SimNet~\cite{Pinheiro2017SimNet} & 69.6 \\
		CDAN-E~\cite{Long2018CDAN} & 70.0 \\
		\hline\hline
		Ours & \textbf{76.2} \\
		\hline\hline
		SE*~\cite{French2018} & 82.8 \\
		\hline
	\end{tabular}
	}
	\label{tab:r50visda_results}
\end{table}

\subsection{DA on Large Datasets}

We apply our method to adaptation on large-scale, large-image datasets. In particular, we evaluate on VisDA-2017~\cite{visda2017} image classification and VisDA-2017 image segmentation tasks.

\paragraph{Classification.}
The VisDA-2017 image classification is a 12-class domain adaptation problem. The source domain consists of 152,397 synthetic images, where 3D CAD models are rendered from various conditions. The target domain consists of 55,388 real images taken from the MS-COCO dataset~\cite{Lin2014COCO}. Since the objective is to learn from labeled synthetic images and correctly predict the class of real images, this dataset has been frequently used in many domain adaptation works~\cite{Long2015DAN, Ganin2016DANN, Saito2018, Saito2018b, French2018}. For fair comparison with recent works, we follow the protocol of ADR~\cite{Saito2018b} in our experiments. Specifically, we apply the \eadd~after the second fully connected layer, and \cadd~within the last convolution layer of ResNet-50~\cite{He2016ResNet} and ResNet-101 models. Both models are initialized with weights from an ImageNet~\cite{Deng2009ImageNet} pre-trained model. For more details on implementation, we refer our readers to Appendix B. 

The per-class adaptation performance with a ResNet-101 backbone can be found in Table~\ref{tab:visda_results}. The table clearly shows that our proposed method surpasses previous methods by a large margin. Note that all methods in this table use the same ResNet-101 backbone. Compared to the performance of a source only model, we achieve a 30.7\% improvement (or 60.4\% relative improvement) on the average accuracy. Furthermore, \methodname\ shows a significant improvement across all categories; in fact, it achieves the best per-class performance in all classes, except the ``truck'' class, where it falls behind ADR by a mere 0.2\%. Although our source only model is slightly lower than that of both MCD~\cite{Saito2018} and ADR, our proposed method effectively generalizes a model from the source to target domain, with stronger adaptation performance over MCD and ADR by margins of 9.6\% and 6.7\%, respectively. 

In Table~\ref{tab:r50visda_results}, we show that it is feasible to apply \methodname\ on a different backbone network with success. Similarly to \methodname\ on ResNet-101, our model outperforms recent previous methods, and demonstrates a significance improvement over the source only model. While SE reports the best overall performance, we do not consider it to be comparable to other methods - including ours - because the reported accuracy is a result of 16 ensembled predictions.


For qualitative analysis, Figure~\ref{fig:tsne} visualizes the feature representations of VisDA-2017 classification with t-SNE~\cite{Maaten2008}. The source only model shows strong clustering of the source domain's synthetic image samples (blue), but fails to have similar influence on the target domain's real image samples (red). During training, \methodname\ constantly enforces the clustering of target samples by stimulating the feature representations and decision boundary of the model. Therefore, we can clearly see an improved separation of target features with \methodname, resulting in the best performance in VisDA-2017.


\paragraph{Segmentation.}\label{ssec:seg}
To further demonstrate our method's applicability to real-world adaptation settings, we evaluate \methodname\ in the challenging VisDA-2017 semantic segmentation task. For the source domain, we use the synthetic GTA5~\cite{Richter2016GTA} dataset which consists of 24966 labeled images. As the target domain, we use the real-world Cityscapes~\cite{Cordts2016Cityscapes}, consisting of 5000 images. Both datasets are evaluated on the same category of 19 classes, with the mean Intersection-over-Union (mIoU) metric. For fair comparison with recent methods~\cite{Ganin2016DANN, Saito2018b}, we follow the procedure of ADR and use a modified version of Fully Convolutional Networks (FCN)~\cite{Long2015FCN} on a ResNet-50 backbone. We apply CAdD within the last convolutional layer of ResNet-50. 

\begin{figure*}[t]
	\centering
	\footnotesize
	\begin{tabular}{c@{\hskip1pt}c@{\hskip1pt}c@{\hskip1pt}c}
		\includegraphics[width=4cm]{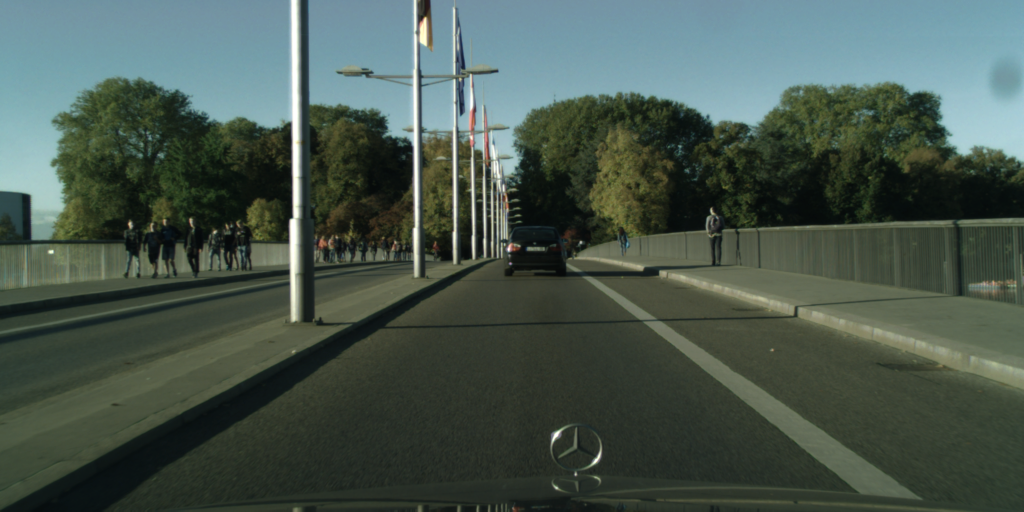}&%
		\includegraphics[width=4cm]{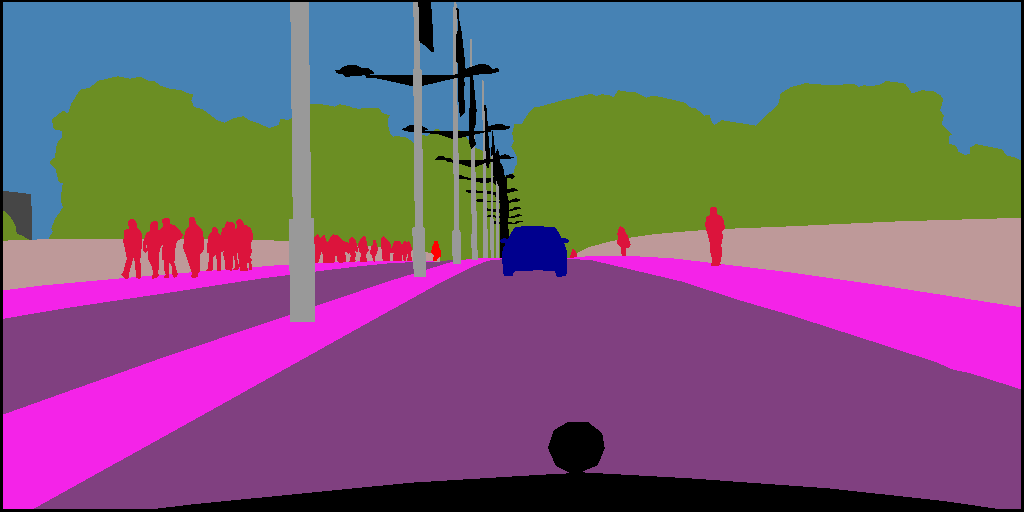}&%
		\includegraphics[width=4cm]{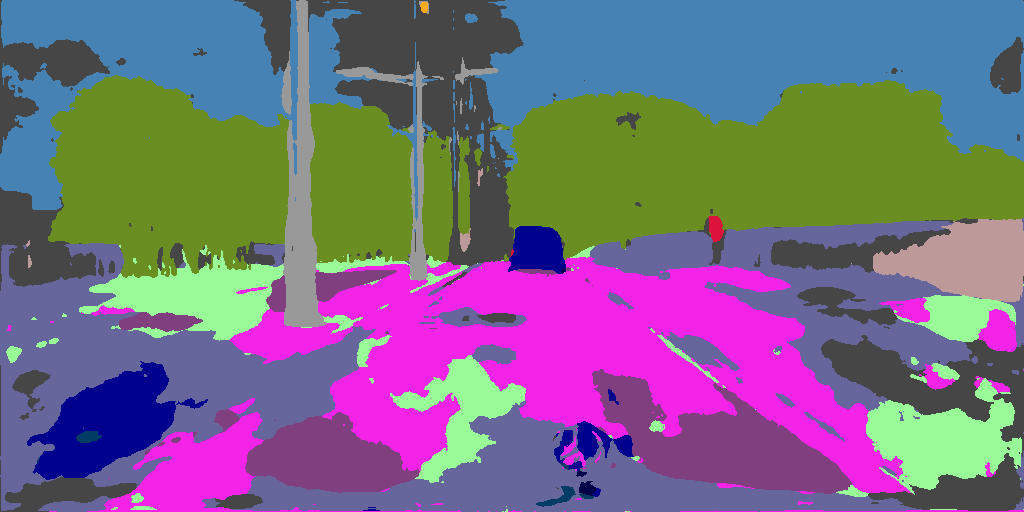}&%
		\includegraphics[width=4cm]{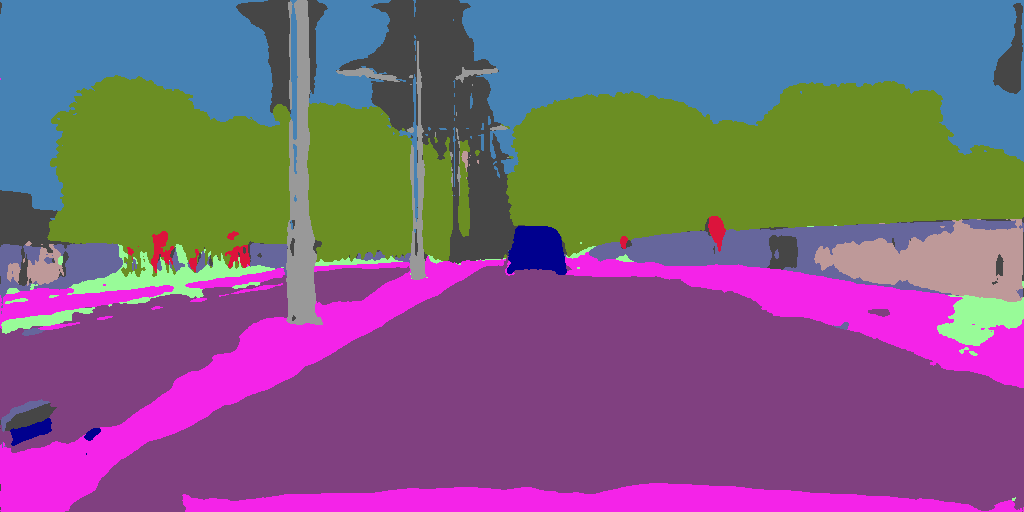}\\
		\includegraphics[width=4cm]{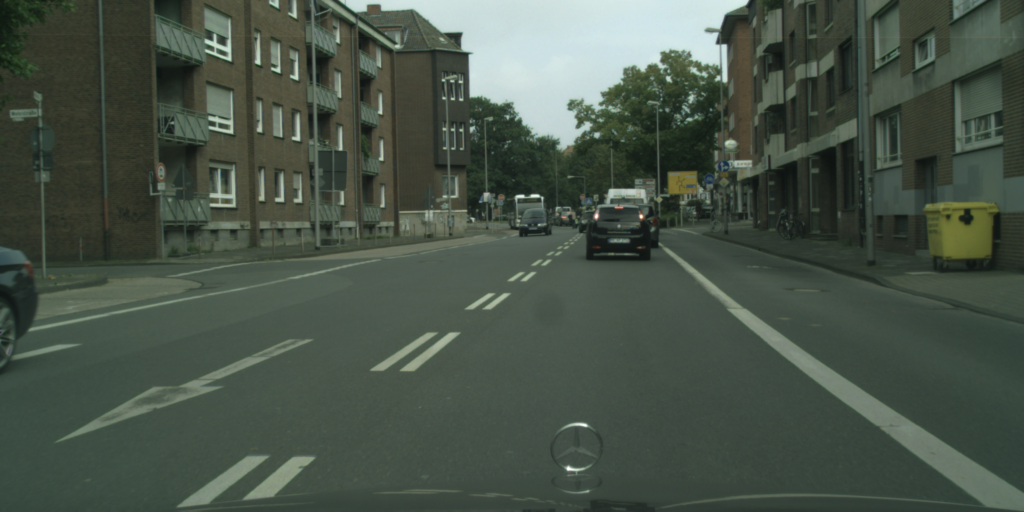}&%
		\includegraphics[width=4cm]{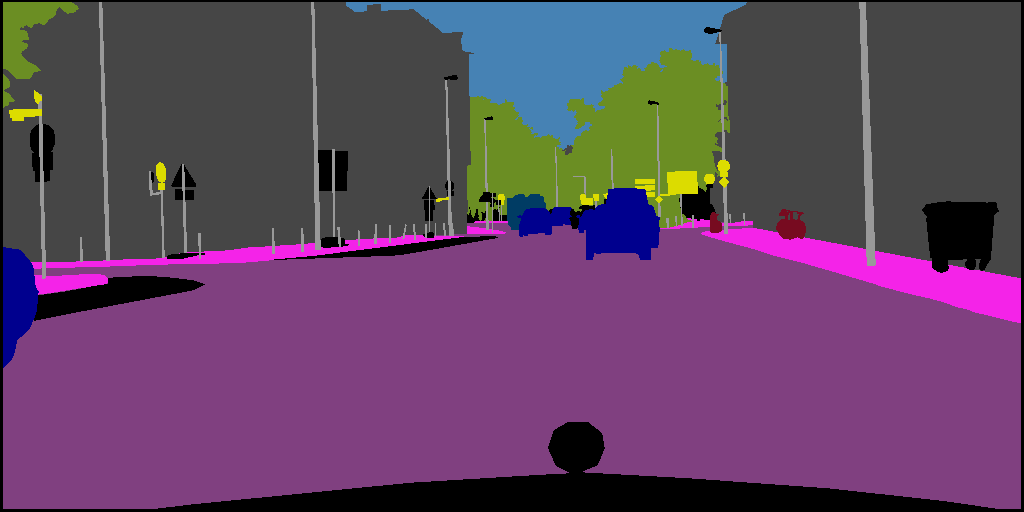}&%
		\includegraphics[width=4cm]{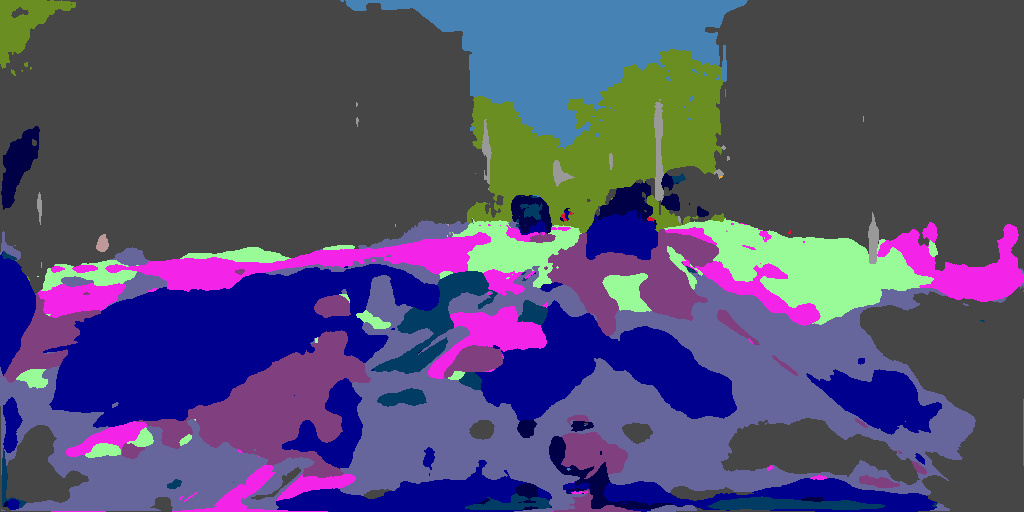}&%
		\includegraphics[width=4cm]{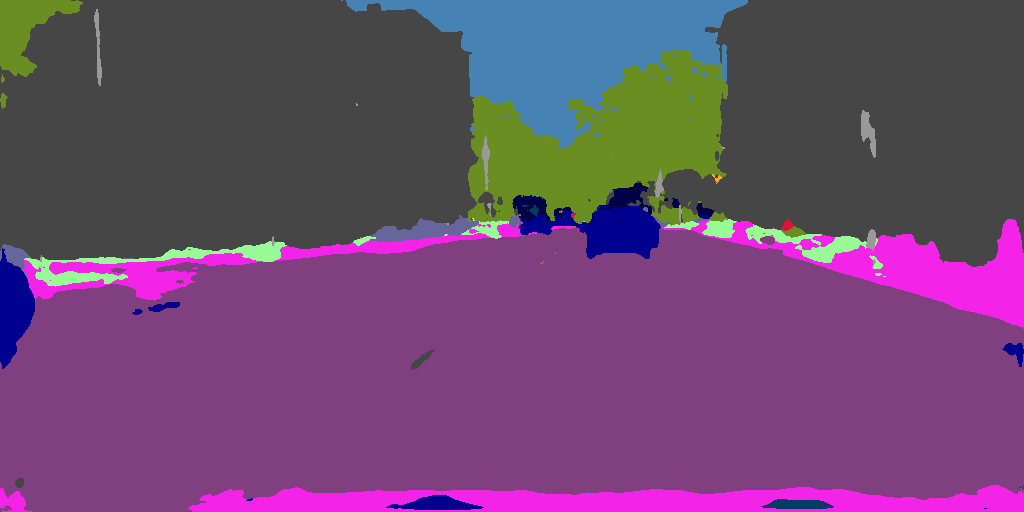}\\
		(a) Input & (b) Ground Truth & (c) Source Only & (d) \methodname \\
	\end{tabular}\vspace{0.3cm}
	\caption{{\bf Semantic segmentation.}\vspace{0.05cm} Qualitative results of the semantic segmentation task on GTA $\rightarrow$ Cityscapes, before and after adaptation with \methodname. We use a modified FCN architecture with ResNet-50 as the base model.}\label{fig:seg}
	\vspace{0.13cm}
\end{figure*}

We report our results in Table~\ref{tab:semantic_results}, alongside results of existing methods. Our method clearly improves upon the mIoU of not only the source only model, but also competing methods. Even with the same training procedure and settings as in the classification experiments, \methodname\ is extremely effective at adapting the most common classes in the dataset. This conclusion is supported in Figure~\ref{fig:seg}, where we display examples of input images, ground truths, and the corresponding outputs of source only and \methodname\ model. While the source only predictions are erroneous in most classes, \methodname's predictions are relatively clean and accurate.
 
\section{Discussion}\label{sec:discussion}
\begin{table*}[t]
	\caption
	{
		Ablation Studies on VisDA-2017 Classification Dataset
	}
	\vspace{0.1cm}
	\centering
	\resizebox{0.9\textwidth}{!}{
		\begin{tabular}{l | c c c c c c c c c c c c | c }	
			\hline\hline
			Methods & aero. & bike & bus & car & horse & knife & moto. & person & plant & sktb. & train & truck & avg. \\
			\hline\hline
			{\bf ResNet-50}&\multicolumn{12}{c|}{}& \\
			Source Only & 54.2 & 27.7 & 17.6 & 57.1 & 48.4 & 4.0 & 86.4 & 11.0 & 69.1 & 15.6 & 95.7 & 7.3 & 46.0 \\
			VAT & 83.1 & 62.5 & 70.5 & 53.0 & 81.8 & 13.2 & 89.9 & 74.4 & 88.5 & 41.1 & 89.0 & 38.2 & 67.1 \\
			fDTA & 88.8 & 58.2 & 82.8 & 82.3 & 90.4 & 0.1 & 92.8 & 77.3 & 94.2 & 78.5 & 86.9 & 0.2 & 72.5 \\
			fDTA + VAT & 91.3 & 66.3 & 77.7 & 77.5 & 91.0 & 13.1 & 92.6 & 83.0 & 94.2 & 58.0 & 85.9 & 12.0 & 73.1 \\
			cDTA & 92.4 & 72.9 & 75.1 & 72.6 & 92.8 & 7.4 & 90.8 & 82.1 & 95.0 & 66.6 & 87.8 & 31.6 & 74.7 \\
			cDTA + VAT & 90.0 & 72.7 & 83.7 & 79.3 & 92.0 & 6.8 & 91.4 & 82.6 & 92.2 & 70.4 & 86.3 & 22.9 & 75.4 \\
			cDTA + fDTA & 88.2 & 68.8 & 87.2 & 82.8 & 92.3 & 5.8 & 89.4 & 78.4 & 95.5 & 74.8 & 82.4 & 16.1 & 75.0 \\
			Ours & 93.1 & 70.5 & 83.8 & 87.0 & 92.3 & 3.3 & 91.9 & 86.4 & 93.1 & 71.0 & 82.0 & 15.3 & 76.2 \\			
			\hline			
			{\bf ResNet-101}&\multicolumn{12}{c|}{}& \\
			Source Only & 46.2 & 27.6 & 31.4 & 78.1 & 71.8 & 1.4 & 71.6 & 14.3 & 63.5 & 31.0 & 93.7 & 3.2 & 50.8 \\
			VAT & 90.1 & 43.9 & 83.9 & 85.6 & 90.9 & 1.4 & 95.0 & 78.6 & 93.8 & 57.9 & 86.2 & 13.4 & 73.2\\
			fDTA & 89.1 & 75.5 & 84.6 & 87.2 & 92.3 & 72.9 & 89.7 & 78.5 & 91.8 & 39.5 & 84.1 & 10.8 & 76.4 \\
			fDTA + VAT & 93.0 & 84.8 & 81.8 & 78.1 & 93.2 & 70.1 & 88.8 & 82.0 & 94.0 & 81.5 & 87.4 & 39.6 & 80.5 \\
			cDTA & 91.8 & 81.5 & 78.7 & 67.0 & 91.3 & 71.6 & 85.3 & 76.9 & 93.5 & 72.5 & 86.7 & 44.1 & 77.0 \\
			cDTA + VAT & 93.8 & 86.1 & 82.9 & 78.3 & 92.2 & 83.9 & 88.2 & 80.6 & 94.1 & 82.2 & 88.0 & 40.0 & 81.2 \\
			cDTA + fDTA & 91.7 & 77.7 & 78.8 & 75.2 & 91.0 & 73.2 & 88.4 & 78.8 & 93.2 & 56.6 & 88.7 & 35.6 & 77.4 \\
			Ours & 93.7 & 82.2 & 85.6 & 83.8 & 93.0 & 81.0 & 90.7 & 82.0 & 95.1 & 78.1 & 86.4 & 32.1 & 81.5 \\
			\hline\hline
	\end{tabular}}
	\label{tab:ablation}
	\vspace{-0.4cm}
\end{table*}

Although the proposed DTA shows significant improvements on multiple visual tasks, we would like to understand the role of each component in \methodname\ and how their combination operates in practice. We perform a series of ablation experiments and present the results in Table~\ref{tab:ablation}. All ablations are conducted on VisDA-2017 image classification dataset. To verify the effectiveness and generality, we use ResNet-50 and ResNet-101 models for all experiments in this ablation. The modified ResNet-based models consist of the original convolutional layers with FAdD after the second fully connected layer, and CAdD within the last convolutional layer. The entropy loss term in Eq.~\eqref{eq:ent} is applied on all ablations except the ``Source Only'' setting.

To assess whether each module of \methodname~(VAT, fDTA, cDTA) plays an important role in the performance, we first experiment with individual modules. Overall, all three modules improve the performance over a source only model. We observe that the three components contribute to the accuracy of each category differently. In ResNet-101, while fDTA has a great impact on the ``knife'' category, VAT significantly boosts the performance of the ``skakteboard'' class. Theoretically, VAT~\cite{Miyato2018} can be seen as the regularization by perturbing the input image, while the proposed methods can be seen as perturbations on the feature space of the model. Therefore, we can see that two combinations (fDTA + VAT), (cDTA + VAT) shows increased performance compared to the individually regularized model (\ie 73.2\%~(VAT) / 77.0\%~(cDTA) $\rightarrow$ 81.2\% in ResNet-101, 67.1\%~(VAT) / 72.5\%~(fDTA) $\rightarrow$ 73.1\% in ResNet-50). These results suggest that it is beneficial to use VAT~\cite{Miyato2018} with the proposed method. More specifically, both methods exhibit complementary effects for adaptation on a large domain shift. This advantage can also be observed in the comparison of fDTA + cDTA to a final version of the proposed method (VAT + fDTA + cDTA). One interesting point is that all these trends are mostly maintained in both backbone models; the only difference is the amount of margin between the performance of source only and individual models. From this fact, we conclude that the proposed method can act as a general regularization technique for adaptation, regardless of the model's capacity. 

\section{Conclusion}\label{sec:conclusions}
We presented a simple yet effective method for unsupervised domain adaptation despite large domain shifts. With two types of proposed adversarial dropout modules, EAdD and CAdD, we enforced the cluster assumption on the target domain. The proposed methods are easily integrated into existing deep learning architectures. Through extensive experiments on various small and large datasets, we demonstrated the effectiveness of the proposed method on two domain adaptation tasks, and in all cases we achieved significant improvement as compared to the source-only model and the state-of-the-art results. 

\vspace{-0.3cm}
\small{
\paragraph*{Acknowledgements.}
This work was supported by Institute for Information \& communications Technology Promotion(IITP) grant funded by the Korea government(MSIT) (No. R7117-16-0164, Development of wide area driving environment awareness and cooperative driving technology which are based on V2X wireless communication).
}

{\small
\bibliographystyle{ieee_fullname}
\bibliography{refs}

\begin{thebibliography}{10}\itemsep=-1pt

\bibitem{Ben2010}
Shai Ben-David, John Blitzer, Koby Crammer, Alex Kulesza, Fernando Pereira, and
  Jennifer~Wortman Vaughan.
\newblock A theory of learning from different domains.
\newblock {\em Mach. Learn.}, 2010.

\bibitem{Ben2006}
Shai Ben-David, John Blitzer, Koby Crammer, and Fernando Pereira.
\newblock Analysis of representations for domain adaptation.
\newblock In {\em NIPS}, 2006.

\bibitem{Bousmalis2017}
Konstantinos Bousmalis, Nathan Silberman, David Dohan, Dumitru Erhan, and Dilip
  Krishnan.
\newblock Unsupervised pixel-level domain adaptation with generative
  adversarial networks.
\newblock In {\em CVPR}, 2017.

\bibitem{Bousmalis2016}
Konstantinos Bousmalis, George Trigeorgis, Nathan Silberman, Dilip Krishnan,
  and Dumitru Erhan.
\newblock Domain separation networks.
\newblock In {\em NIPS}, 2016.

\bibitem{Chapelle2005}
Olivier Chapelle and Alexander Zien.
\newblock Semi-supervised classification by low density separation.
\newblock In {\em AISTATS}, 2005.

\bibitem{Coates2011}
Adam Coates, Honglak Lee, and Andrew Ng.
\newblock An analysis of single layer networks in unsupervised feature
  learning.
\newblock In {\em AISTATS}, 2011.

\bibitem{Cordts2016Cityscapes}
Marius Cordts, Mohamed Omran, Sebastian Ramos, Timo Rehfeld, Markus Enzweiler,
  Rodrigo Benenson, Uwe Franke, Stefan Roth, and Bernt Schiele.
\newblock The cityscapes dataset for semantic urban scene understanding.
\newblock In {\em CVPR}, 2016.

\bibitem{Deng2009ImageNet}
Jia Deng, Wei Dong, Richard Socher, Li-Jia Li, Kai Li, and Li Fei-Fei.
\newblock Imagenet: A large-scale hierarchical image database.
\newblock In {\em CVPR}, 2009.

\bibitem{French2018}
Geoffrey French, Michal Mackiewicz, and Mark Fisher.
\newblock Self-ensembling for visual domain adaptation.
\newblock In {\em ICLR}, 2018.

\bibitem{VirtualKITTI2016}
A Gaidon, Q Wang, Y Cabon, and E Vig.
\newblock Virtual worlds as proxy for multi-object tracking analysis.
\newblock In {\em CVPR}, 2016.

\bibitem{Ganin2015}
Yaroslav Ganin and Victor Lempitsky.
\newblock Unsupervised domain adaptation by backpropagation.
\newblock In {\em ICML}, 2015.

\bibitem{Ganin2016DANN}
Yaroslav Ganin, Evgeniya Ustinova, Hana Ajakan, Pascal Germain, Hugo
  Larochelle, Fran\c{c}ois Laviolette, Mario Marchand, and Victor Lempitsky.
\newblock Domain-adversarial training of neural networks.
\newblock {\em JMLR}, 2016.

\bibitem{Geiger2013KITTI}
Andreas Geiger, Philip Lenz, Christoph Stiller, and Raquel Urtasun.
\newblock Vision meets robotics: The kitti dataset.
\newblock {\em IJRR}, 2013.

\bibitem{He2016ResNet}
Kaiming He, Xiangyu Zhang, Shaoqing Ren, and Jian Sun.
\newblock Deep residual learning for image recognition.
\newblock In {\em CVPR}, 2016.

\bibitem{Hoffman2018}
Judy Hoffman, Eric Tzeng, Taesung Park, Jun-Yan Zhu, Phillip Isola, Kate
  Saenko, Alexei~A. Efros, and Trevor Darrell.
\newblock Cy{CADA}: {C}ycle-consistent adversarial domain adaptation.
\newblock In {\em ICML}, 2018.

\bibitem{Hou2019WCD}
Saihui Hou and Zilei Wang.
\newblock Weighted channel dropout for regularization of deep convolutional
  neural network.
\newblock In {\em AAAI}, 2019.

\bibitem{Hull1994}
Jonathan~J. Hull.
\newblock A database for handwritten text recognition research.
\newblock {\em IEEE TPAMI}, 1994.

\bibitem{Kingma2015}
Diedrik~P. Kingma and Jimmy Ba.
\newblock Adam: {A} method for stochastic optimization.
\newblock In {\em ICLR}, 2015.

\bibitem{Krizhevsky2009}
Alex Krizhevsky.
\newblock Learning multiple layers of features from tiny images, 2009.
\newblock {\em Tech Report}.

\bibitem{CorregulNIPS2018}
Abhishek Kumar, Prasanna Sattigeri, Kahini Wadhawan, Leonid Karlinsky, Rogerio
  Feris, Bill Freeman, and Gregory Wornell.
\newblock Co-regularized alignment for unsupervised domain adaptation.
\newblock In {\em NIPS}, 2018.

\bibitem{Lecun1998}
Yann LeCun, Leon Bottou, Yoshua Bengio, and Pattrick Haffner.
\newblock Gradient-based learning applied to document recognition.
\newblock {\em Proc. IEEE}, 1998.

\bibitem{Lin2014COCO}
Tsung-Yi Lin, Michael Maire, Serge Belongie, Lubomir Bourdev, Ross Girshick,
  James Hays, Pietro Perona, Deva Ramanan, C.~Lawrence Zitnick, and Piotr
  Doll\'{a}r.
\newblock Microsoft coco: Common objects in context.
\newblock In {\em ECCV}, 2014.

\bibitem{Long2015FCN}
Jonathan Long, Evan Shelhamer, and Trevor Darrell.
\newblock Fully convolutional networks for semantic segmentation.
\newblock In {\em CVPR}, 2015.

\bibitem{Long2015DAN}
Mingsheng Long, Yue Cao, Jianmin Wang, and Michael~I. Jordan.
\newblock Learning transferable features with deep adaptation networks.
\newblock In {\em ICML}, 2015.

\bibitem{Long2018CDAN}
Mingsheng Long, Zhangjie Cao, Jianmin Wang, and Michael~I. Jordan.
\newblock Conditional adversarial domain adaptation.
\newblock In {\em NIPS}, 2018.

\bibitem{Long2016RTN}
Mingsheng Long, H. Zhu, J. Wang, and Michael~I. Jordan.
\newblock Unsupervised domain adaptation with residual transfer networks.
\newblock In {\em NIPS}, 2016.

\bibitem{Long2017JAN}
Mingsheng Long, Han Zhu, Jianmin Wang, and Michael~I. Jordan.
\newblock Deep transfer learning with joint adaptation networks.
\newblock In {\em ICML}, 2017.

\bibitem{Miyato2018}
T. Miyato, S. i. Maeda, M. Koyama, and S. Ishii.
\newblock Virtual adversarial training: a regularization method for supervised
  and semi-supervised learning.
\newblock {\em IEEE TPAMI}, 2018.

\bibitem{Netzer2011}
Yuval Netzer, Tao Wang, Adam Coates, Alessandro Bissacco, Bo Wu, and Andrew~Y.
  Ng.
\newblock Reading digits in natural images with unsupervised feature learning.
\newblock In {\em NIPS workshop on deep learning and unsupervised feature
  learning}, 2011.

\bibitem{Park2014Dropout}
Sungheon Park and Nojun Kwak.
\newblock Analysis on the dropout effect in convolutional neural networks.
\newblock {\em ACCV}, 2016.

\bibitem{Park2018}
Sungrae Park, JunKeon Park, Su-Jin Shin, and Il-Chul Moon.
\newblock Adversarial dropout for supervised and semi-supervised learning.
\newblock In {\em AAAI}, 2018.

\bibitem{Pei2018MADA}
Zhongyi Pei, Zhangjie Cao, Mingsheng Long, and Jianmin Wang.
\newblock Multi-adversarial domain adaptation.
\newblock In {\em AAAI}, 2018.

\bibitem{visda2017}
Xingchao Peng, Ben Usman, Neela Kaushik, Judy Hoffman, Dequan Wang, and Kate
  Saenko.
\newblock {VisDA}: The visual domain adaptation challenge, 2017.

\bibitem{Pinheiro2017SimNet}
Pedro~O. Pinheiro.
\newblock Unsupervised domain adaptation with similarity learning.
\newblock In {\em CVPR}, 2018.

\bibitem{Richter2016GTA}
Stephan~R. Richter, Vibhav Vineet, Stefan Roth, and Vladlen Koltun.
\newblock Playing for data: {G}round truth from computer games.
\newblock In {\em ECCV}, 2016.

\bibitem{SynthiaData}
German Ros, Laura Sellart, Joanna Materzynska, David Vazquez, and Antonio
  Lopez.
\newblock {The SYNTHIA Dataset}: A large collection of synthetic images for
  semantic segmentation of urban scenes.
\newblock In {\em CVPR}, 2016.

\bibitem{Saito2018b}
Kuniaki Saito, Yoshitaka Ushiku, Tatsuya Harada, and Kate Saenko.
\newblock Adversarial dropout regularization.
\newblock In {\em ICLR}, 2018.

\bibitem{Saito2018}
Kuniaki Saito, Kohei Watanabe, Yoshitaka Ushiku, and Tatsuya Harada.
\newblock Maximum classifier discrepancy for unsupervised domain adaptation.
\newblock In {\em CVPR}, 2018.

\bibitem{Sank2018GTA}
Swami Sankaranarayanan, Yogesh Balaji, Carlos~D. Castillo, and Rama Chellappa.
\newblock Generate to adapt: Aligning domains using generative adversarial
  networks.
\newblock In {\em CVPR}, 2018.

\bibitem{Selvaraju_2017_ICCV}
Ramprasaath~R. Selvaraju, Michael Cogswell, Abhishek Das, Ramakrishna Vedantam,
  Devi Parikh, and Dhruv Batra.
\newblock Grad-cam: Visual explanations from deep networks via gradient-based
  localization.
\newblock In {\em IEEE ICCV}, 2017.

\bibitem{Shu2018}
Rui Shu, Hung Bui, Hirokazu Narui, and Stefano Ermon.
\newblock {A DIRT-T} approach to unsupervised domain adaptation.
\newblock In {\em ICLR}, 2018.

\bibitem{Srivastava2014Dropout}
Nitish Srivastava, Geoffrey Hinton, Alex Krizhevsky, Ilya Sutskever, and Ruslan
  Salakhutdinov.
\newblock Dropout: A simple way to prevent neural networks from overfitting.
\newblock {\em JMLR}, 2014.

\bibitem{Taigman2017DTN}
Yaniv Taigman, Adam Polyak, and Lior Wolf.
\newblock Unsupervised cross-domain image generation.
\newblock {\em ICLR}, 2017.

\bibitem{Tompson2015SpatialDropout}
Jonathan Tompson, Ross Goroshin, Arjun Jain, Yann LeCun, and Christoph Bregler.
\newblock Efficient object localization using convolutional networks.
\newblock In {\em CVPR}, 2015.

\bibitem{Tzeng2017}
Eric Tzeng, Judy Hoffman, Kate Saenko, and Trevor Darrell.
\newblock Adversarial discriminative domain adaptation.
\newblock In {\em CVPR}, 2017.

\bibitem{Maaten2008}
Laurens van~der Maaten and G. Hinton.
\newblock Visualizing data using t-sne.
\newblock {\em Journal of Machine Learning Research}, 2008.

\bibitem{Volpi2018AFA}
Riccardo Volpi, Pietro Morerio, Silvio Savarese, and Vittorio Murino.
\newblock Adversarial feature augmentation for unsupervised domain adaptation.
\newblock In {\em CVPR}, 2018.

\bibitem{Wang2019TADA}
Ximei Wang, Liang Li, Weirui Ye, Mingsheng Long, and Jianmin Wang.
\newblock Transferable attention for domain adaptation.
\newblock In {\em AAAI}, 2019.

\bibitem{Yosinski2014TransfeFea}
Jason Yosinski, Jeff Clune, Yoshua Bengio, and Hod Lipson.
\newblock How transferable are features in deep neural networks?
\newblock In {\em NIPS}, 2014.

\bibitem{Zeiler2014VisCNN}
Matthew~D. Zeiler and Rob Fergus.
\newblock Visualizing and understanding convolutional networks.
\newblock In {\em ECCV}, 2014.

\bibitem{Zhang2016FineGrained}
Xiaopeng Zhang, Hongkai Xiong, Wengang Zhou, Weiyao Lin, and Qi Tian.
\newblock Picking deep filter responses for fine-grained image recognition.
\newblock In {\em CVPR}, 2016.

\bibitem{Zhu2017}
Jun-Yan. Zhu, Taesung Park, Phillip Isola, and Alexei~A. Efros.
\newblock Unpaired image-to-image translation using cycle-consistent
  adversarial networks.
\newblock In {\em ICCV}, 2017.

\end{thebibliography}


\begin{thebibliography}{1}\itemsep=-1pt

\bibitem{Kingma2015}
Diedrik~P. Kingma and Jimmy Ba.
\newblock Adam: {A} method for stochastic optimization.
\newblock In {\em ICLR}, 2015.

\bibitem{Long2015FCN}
Jonathan Long, Evan Shelhamer, and Trevor Darrell.
\newblock Fully convolutional networks for semantic segmentation.
\newblock In {\em CVPR}, 2015.

\bibitem{Park2018}
Sungrae Park, JunKeon Park, Su-Jin Shin, and Il-Chul Moon.
\newblock Adversarial dropout for supervised and semi-supervised learning.
\newblock In {\em AAAI}, 2018.

\bibitem{Selvaraju_2017_ICCV}
Ramprasaath~R. Selvaraju, Michael Cogswell, Abhishek Das, Ramakrishna Vedantam,
  Devi Parikh, and Dhruv Batra.
\newblock Grad-cam: Visual explanations from deep networks via gradient-based
  localization.
\newblock In {\em IEEE ICCV}, 2017.

\bibitem{Shu2018}
Rui Shu, Hung Bui, Hirokazu Narui, and Stefano Ermon.
\newblock {A DIRT-T} approach to unsupervised domain adaptation.
\newblock In {\em ICLR}, 2018.

\end{thebibliography}
}

\onecolumn
\pagebreak

\begin{appendices}
\maketitle
\renewcommand{\thepage}{A-\arabic{page}}
\renewcommand{\thesection}{Appendix \Alph{section}}
\renewcommand{\thetable}{A-\arabic{table}}
\renewcommand{\thefigure}{A-\arabic{figure}}
\renewcommand{\theequation}{a-\arabic{equation}}

\setcounter{figure}{0}
\setcounter{table}{0}
\setcounter{section}{0}

We derive an approximation of the channel-wise adversarial dropout ($\S$~\ref{sec:approx}) and provide implementation details of the experiments ($\S$~\ref{sec:impl}). Lastly, we provide additional GradCAM visualizations ($\S$~\ref{sec:gradcam_vis}).
\section{Approximation of Channel-wise Adversarial Dropout}\label{sec:approx}
Without loss of generality, the dropout mask $\dropmask$ is vectorized to $\vmask=vec(\dropmask) \in \mathbb{R}^{CHW}. $
Similarly, $\vmask^0$ and $\vmask^s$ represent vectorized forms of $\dropmask^0$ and $\dropmask^s$, respectively. After vectorization of $\dropmask$, we refer to the elements of $\dropmask(i)$ with a set of indices $\pi_i$, and impose the channel-wise dropout constraints as follows:
\begin{equation}
	\vmask[\pi_i] = vec(\dropmask(i)) = \zero\ \text{or}\ \one \in \mathbb{R}^{HW}.\label{eq:vector_constraint}
\end{equation}

Let denote $d(\bx{}, \vmask; \vmask^s) = D\left[h(\bx{};\vmask^s), h(\bx{};\vmask)\right]$  as the divergence between two outputs using different dropout masks for convenience sake. Assuming $d$ is a differentiable function with respect to $\vmask$, it can be approximated by a first-order Taylor expansion:
\begin{gather}
	d(\bx{}, \vmask; \vmask^s) \approx d(\bx{}, \vmask^0; \vmask^s) + (\vmask - \vmask^0)^T \jacobian
	~\text{where $\jacobian = \nabla_{\vmask} d(\bx{}, \vmask; \vmask^s)\Bigr|_{\vmask = \vmask^0}$}.\nonumber
\end{gather}
This equation shows that the Jacobian is proportional to the divergence. In other words,
\begin{align}
\label{eq:proportional}
d(\bx{}, \vmask; \vmask^s) \propto \vmask^T \jacobian.
\end{align}
We now see that the elements of $\jacobian$ correspond to the \textit{impact values}, which indicate the contribution of each activation over the divergence metric. Thus, for the given Jacobian, we can systematically modify the elements of $\vmask$ to maximize the divergence. However, due to the channel-wise dropout constraint from Eq.~\eqref{eq:vector_constraint}, we cannot modify each element individually. Instead, we reformulate the above relationship as:
\begin{align}
d(\bx{}, \vmask; \vmask^s) \propto \sum_{i}^{C} \vmask[\pi_i]^T \jacobian[\pi_i].
\end{align}
The impact value $\boldsymbol{s}$ of the $i$-th activation map in $\lowernet(\bx{})$ can be defined as:
\begin{align}
\boldsymbol{s}_i = \one^T\jacobian[\pi_i],\label{jacosum_def}
\end{align}
Consequently, after computing the impact values $\boldsymbol{s}$, we solve 0/1 Knapsack problem as proposed in~\cite{Park2018} while holding the constraints~\eqref{eq:vector_constraint}.

\section{Implementation Details}\label{sec:impl}
\subsection*{Training with DTA Loss} 
We apply a ramp-up factor on DTA loss function $\lossfunc{DTA}$ to stabilize the training process. Instead of directly modulating the weight term $\lambda_1$, we gradually increase the perturbation magnitudes $\delta_e$ and $\delta_c$ which decide the number of hidden units to be eliminated. It allows us to regulate the consistency term, and to train the network being robust to various levels of perturbation generated by the adversarial dropout. We update  the ramp-up factors with the following schedule:
\begin{equation}
\beta^{(t)} = \min(1, \tfrac{t}{T_r}),
\end{equation}
where  $T_r$ represents the ramp-up period, and $\beta^{(t)}$ denotes the ramp up factor at the current epoch $t$. Finally, the perturbation magnitude is defined as: 
\begin{equation}
\delta^{(t)} = \beta^{(t)}\bar{\delta},
\end{equation}
where $\bar{\delta}$ denotes the maximum level of perturbation. In practice, the same ramp-up period $T_r$ is applied for both $\delta_e$ and $\delta_c$.

\subsection*{Hyperparameters}
Table~\ref{tab:hyperparams} presents the hyperparameters used in our experiments. We followed a similar hyperparameter search protocol as Shu \etal~\cite{Shu2018}, where we sample a very small subset of labels from the target domain training set. For each objective function, we limit the hyperparameter search to a predefined set of values: $\lambda_1=\{2\}$, $\lambda_2=\{0, 0.01, 0.02\}$, $\lambda_3=\{0, 0.1, 0.2\}$, $\delta_e=\{0, 0.1\}$, $\delta_c=\{0, 0.01, 0.02, 0.05\}$, and $\epsilon=\{0, 3.5, 15\}$. Furthermore, we provide the rest of parameters related to network training for each experimental set up. 

\paragraph{Small dataset.} 
All small dataset experiments were trained for 90 epochs, using Adam optimizer~\cite{Kingma2015} with an initial learning rate of 0.001, decaying by a factor of 0.1 every 30 epochs. 

\paragraph{Large dataset.}
We conducted the VisDA-2017 classification experiments on ResNet-50 and ResNet-101. We trained the networks for 20 epochs using Stochastic Gradient Descent (SGD) with a momentum value of 0.9 and an initial learning rate of 0.001, which decays by a factor of 0.1 after 10th epoch. 

\paragraph{Semantic segmentation.}
The semantic segmentation task for domain adaptation from GTA5 to Cityscapes was trained for 5 epochs using SGD with a momentum of 0.9. Since FCN~\cite{Long2015FCN} has no fully-connected layers, $\bar{\delta_e}$ was automatically set to 0. In addition, we used the maximum $\bar{\delta_c}$ value from the beginning because the task-specific objective were dominant in the early stages of training. In this experiment, we turned off VAT objective which hinders from learning the segmentation task.

\begin{table*}
	\centering
	\caption{Hyperparameters}
	\label{tab:hyperparams}
	\begin{tabular}{lcccccccc}
		\hline\hline\\[-1em]
		Experiment & Backbone  & $\lambda_1$ & $\lambda_2$ & $\lambda_3$ & $\bar{\delta_e}$ & $\bar{\delta_c}$ & $T_r$ & $\epsilon$ \\ 	
		\hline\\[-0.7em]
		\multicolumn{9}{l}{{\bf Small dataset}}\\[0.1em]
		SVHN $\rightarrow$ MNIST &9 Conv+1 FC & 2 & 0.01 & 0.1 & 0.1 & 0.05 & 80 & 3.5 \\
		MNIST $\rightarrow$ USPS & 3 Conv+2 FC & 2 & 0.01 & 0 & 0.1 & 0.05 & 80 & 0 \\
		USPS $\rightarrow$ MNIST & 3 Conv+2 FC & 2 & 0.01 & 0.1 & 0.1 & 0.05 & 80 & 3.5 \\
		STL $\rightarrow$ CIFAR     & 9 Conv+1 FC & 2 & 0.01 & 0.1 & 0 & 0.05 & 60 & 3.5 \\
		CIFAR $\rightarrow$ STL     &  9 Conv+1 FC & 2 & 0.01 & 0 & 0 & 0.05 & 80 & 0 \\
		\hline\\[-0.7em]
		\multicolumn{9}{l}{{\bf Large dataset}}\\[0.1em]
		VisDA-2017 Classification & ResNet-50 & 2 & 0.02 & 0.2 & 0.1 & 0.01 & 20 & 15 \\
		VisDA-2017 Classification & ResNet-101 & 2 & 0.02 & 0.2 & 0.1 & 0.01 & 30 & 15 \\
		\hline	\\[-0.7em]
		\multicolumn{9}{l}{{\bf Semantic segmentation}}\\[0.1em]
		GTA5 $\rightarrow$ Cityscapes & ResNet-50 FCN & 2 & 0.01 & 0 & 0 & 0.02 & 1 & 0 \\	
		\hline\hline
	\end{tabular}
\end{table*}

\newpage
\section{Additional GradCAM visualizations}\label{sec:gradcam_vis}

\begin{figure}[h]
	\centering
	\footnotesize
	\begin{tabular}{c@{\hskip1pt}c@{\hskip1pt}c@{\hskip1pt}c@{\hskip-.5pt}c}
		\includegraphics[width=3cm]{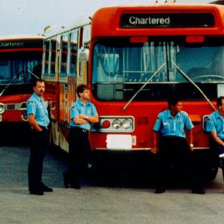}&%
		\includegraphics[width=3cm]{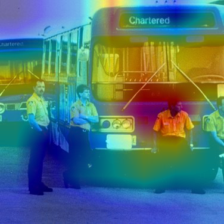}&%
		\includegraphics[width=3cm]{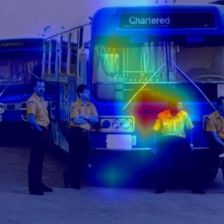}&%
		\includegraphics[width=3cm]{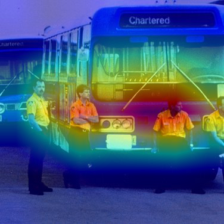}&%
		\includegraphics[width=3cm]{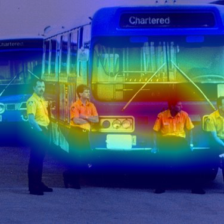}\\[-0.2em]
		\includegraphics[width=3cm]{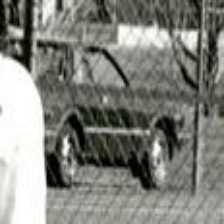}&%
		\includegraphics[width=3cm]{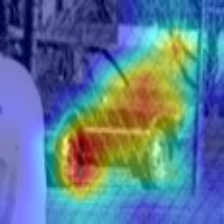}&%
		\includegraphics[width=3cm]{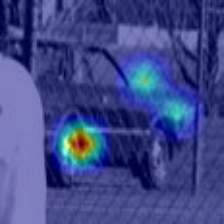}&%
		\includegraphics[width=3cm]{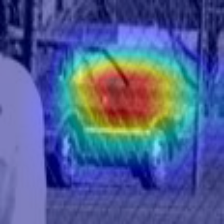}&%
		\includegraphics[width=3cm]{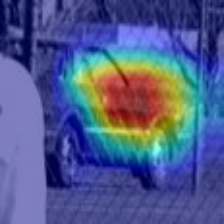}\\[-0.2em]
		\includegraphics[width=3cm]{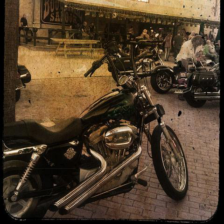}&%
		\includegraphics[width=3cm]{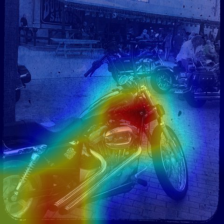}&%
		\includegraphics[width=3cm]{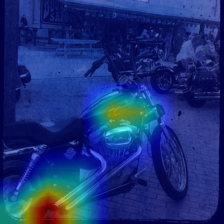}&%
		\includegraphics[width=3cm]{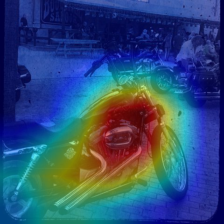}&%
		\includegraphics[width=3cm]{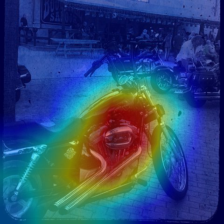}\\[-0.2em]
		\includegraphics[width=3cm]{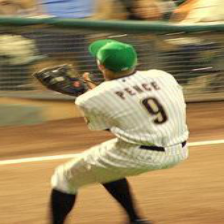}&%
		\includegraphics[width=3cm]{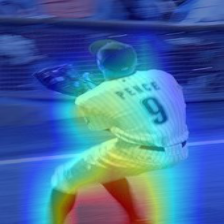}&%
		\includegraphics[width=3cm]{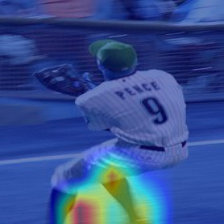}&%
		\includegraphics[width=3cm]{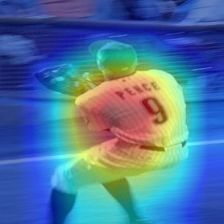}&%
		\includegraphics[width=3cm]{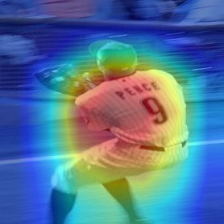}\\[-0.2em]
		\includegraphics[width=3cm]{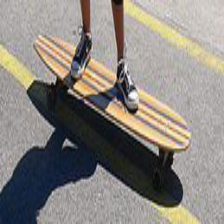}&%
		\includegraphics[width=3cm]{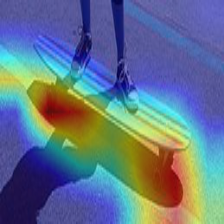}&%
		\includegraphics[width=3cm]{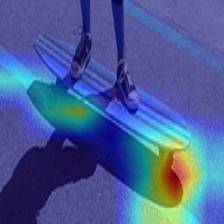}&%
		\includegraphics[width=3cm]{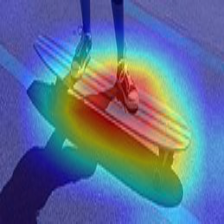}&%
		\includegraphics[width=3cm]{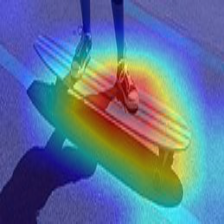}\\[-0.2em]
		(a) Input & (b) SO & (c) SO+AdD  & (d) DTA & (e) DTA+AdD\\
	\end{tabular}\vspace{0.2cm}
	\caption{ Effect of adversarial dropout, visualized by GradCAM~\cite{Selvaraju_2017_ICCV}.} 
	\label{fig:gradcam_additional}
\end{figure}

In Figure~\ref{fig:gradcam_additional}, we provide additional GradCAM visualizations to highlight the effects of adversarial dropout. 

\end{appendices}

\end{document}